\documentclass{article} 
\usepackage{iclr2023_conference,times}

\usepackage{amsmath,amsfonts,bm}

\def\eqref#1{equation~\ref{#1}}

\def\1{\bm{1}}

\usepackage{hyperref}
\usepackage{url}
\usepackage{amsmath}
\usepackage{algorithmic}
\usepackage{array}

\usepackage{fixltx2e}

\usepackage{xcolor}
\iftrue
\newcommand{\berivan}[1]{{\color{red}\textbf{BI}: #1}}

\fi
\usepackage{url}
\usepackage{graphicx}
\usepackage{subfigure}
\usepackage{url}
\usepackage{multicol}
\usepackage{multirow}
\usepackage{wrapfig}
\usepackage{float}
\usepackage{tabularx}
\usepackage{caption}
\usepackage{enumitem}
\usepackage{soul}

\usepackage{array}
\newcolumntype{P}[1]{>{\centering\arraybackslash}p{#1}}
\newcommand{\centered}[1]{\begin{tabular}{c} #1 \end{tabular}}
\usepackage{amsthm}

\newtheorem{definition}{Definition}
\usepackage{tabularx}
\usepackage{booktabs}    

\usepackage{amsmath,amssymb,amsfonts}
\usepackage{algorithmic}
\usepackage{graphicx}
\usepackage{textcomp}
\usepackage{xcolor}
\usepackage{subfigure}
\usepackage[numberedsection=false]{glossaries}
\usepackage{amsthm}
\usepackage{dblfloatfix}

\newtheorem{remark}{\bf Remark}[section]

\newacronym{iot}{IoT}{Internet of Things}
\newacronym{ml}{ML}{machine learning}
\newacronym{dl}{DL}{Deep Learning}
\newacronym{marl}{MARL}{multi-agent reinforcement learning}
\newacronym{rl}{RL}{reinforcement learning}
\newacronym{decpomdp}{Dec-POMDP}{Decentralized Partially Observable Markov Decision Process }
\newacronym{pomdp}{POMDP}{Partially Observable Markov Decision Process}
\newacronym{uav}{UAV}{Unmanned Aerial Vehicle}
\newacronym{dqn}{DQN}{Deep Q-Network}
\newacronym{dnn}{DNN}{deep neural network}
\newacronym{dial}{DIAL}{Differentiable Inter-Agent Learning}
\newacronym{mdp}{MDP}{Markov decision process}
\newacronym{fov}{FoV}{Field of View}
\newacronym{cnn}{CNN}{Convolutional Neural Network}
\newacronym{nn}{NN}{neural network}
\newacronym{ddql}{DDQL}{Distributed Deep Q-Learning}
\newacronym{pdf}{PDF}{Probability Density Function}
\newacronym{ndpomdp}{ND-POMDP}{Networked Distributed Partially Observable Markov Decision Process}
\newacronym{radam}{RAdam}{Rectified Adam}
\newacronym{cdf}{CDF}{cumulative distribution function}
\newacronym{mpc}{MPC}{Model Predictive Control}
\newacronym{rv}{rv}{Random Variable}
\newacronym{qoe}{QoE}{Quality of Experience}
\newacronym{tlc}{TLC}{Telecommunications}
\newacronym{cml}{CML}{communications for machine learning}
\newacronym{mlc}{MLC}{machine learning for communications}
\newacronym{drl}{DRL}{deep reinforcement rearning}
\newacronym{rf}{RF}{Radio Frequency}
\newacronym{urllc}{URLLC}{Ultra-Reliable and Low-Latency Communications}
\newacronym{fl}{FL}{federated learning}
\newacronym{kpi}{KPI}{Key Performance Indicators}
\newacronym{mec}{MEC}{Mobile Edge Computing}
\newacronym{ei}{EI}{Edge Intelligence}
\newacronym{bs}{BS}{base station}
\newacronym{sdn}{SDN}{Software Defined Networking}
\newacronym{mimo}{MIMO}{Multiple-Input Multiple-Output}
\newacronym{gp}{GP}{Gaussian Process}
\newacronym{iiot}{IIoT}{Industrial Internet of Things}
\newacronym{csi}{CSI}{Channel State Information}
\newacronym{sgd}{SGD}{Stochastic Gradient Descent}
\newacronym{iid}{i.i.d.}{independent and identically distributed}
\newacronym{ofdm}{OFDM}{Orthogonal Frequency Division Multiplexing}
\newacronym{los}{LOS}{Line-of-Sight}
\newacronym{nlos}{NLOS}{Non-Line-of-Sight}
\newacronym{snr}{SNR}{Signal to Noise Ratio}
\newacronym{rb}{RB}{Resource Block}
\newacronym{6g}{6G}{sixth generation}
\newacronym{ai}{AI}{artificial intelligence}
\newacronym{sfl}{SFL}{Synchronous Federated Learning}
\newacronym{frfl}{FRFL}{Fixed Rate Federated Learning}
\newacronym{pgm}{PGM}{Probabilistic Graphical Model}
\newacronym{hmm}{HMM}{Hidden Markov Model}
\newacronym{elbo}{ELBO}{Evidence Lower Bound}
\newacronym{pmf}{PMF}{Probability Mass Function}
\newacronym{smab}{SMAB}{Stochastic Multi-Armed Bandit}
\newacronym{mab}{MAB}{multi-armed bandit}
\newacronym{mc}{MC}{Monte Carlo}
\newacronym{is}{IS}{Importance Sampling}
\newacronym{dms}{DMS}{discrete memoryless source}
\newacronym{ucb}{UCB}{upper confidence bound}
\newacronym{ser}{SER}{Symbol Error Rate}
\newacronym{sc}{SC}{Semantic Communications}
\newacronym{voi}{VoI}{Value of Information}
\newacronym{nlp}{NLP}{natural language processing}
\newacronym{ts}{TS}{Thompson Sampling}
\newacronym{cmab}{CMAB}{contextual multi-armed bandit}
\newacronym{rccmab}{RC-CMAB}{rate-constrained \gls{cmab}}
\newacronym{rcmab}{R-CMAB}{remote \gls{cmab}}
\newacronym{ib}{IB}{information bottleneck}
\newacronym{merl}{MERL}{maximum entropy reinforcement learning}
\newacronym{fedpm}{$\mathtt{FedPM}$}{Federated Probabilistic Mask Training}
\newacronym{lth}{LTH}{Lottery Ticket Hypothesis}
\newacronym{dp}{DP}{differential privacy}
\usepackage{hyperref}
\usepackage{algorithmic}
\usepackage{algorithm}
\usepackage{url}

\newcommand{\E}{\mathbb{E}}

\title{Communication-Efficient Federated Sub-Network Training}
\title{Finding Federated Sub-Networks: A Communication-Efficient Training Strategy}
\title{Federated Supermasks: A Communication-Efficient Federated Learning Strategy}
\title{Federated Finding of Sub-Networks in Randomly Weighted Neural Networks}
\title{Federated Supermasks: Improving Communication Efficiency of Federated Learning}
\title{Federated Supermasks: \\ Communication-Efficient Federated Learning}
\title{Federated Binary Masks: \\ 1-Bit Federated Sub-Network Finding}
\title{1-Bit Federated Sub-Network Finding}
\title{Randomly Weighted Sparse Networks for Communication-Efficient Private Federated Learning}
\title{Randomly Weighted Sub-Networks for Communication-Efficient Private Federated Learning}
\title{Randomly Weighted Sparse Networks for Bitrate-Efficient Private Federated Learning}
\title{Randomly Weighted Sparse Networks for \\ Communication-Efficient Federated Learning}
\title{Sparse Networks with Random Weights for \\ Communication-Efficient Federated Learning}
\title{Sparse Random Networks for \\ Communication-Efficient Federated Learning}

\author{Berivan Isik$^{\mathparagraph}$\thanks{First two authors contributed equally. Work done while F.P. was visiting Imperial College London.} , \ Francesco Pase$^{\mathsection}$$^{*}$, \ Deniz Gunduz$^{\ddagger}$, \  Tsachy Weissman$^{\mathparagraph}$, \  Michele Zorzi$^{\mathsection}$\\
$^{\mathparagraph}$Stanford University, $^{\mathsection}$University of Padova, $^{\ddagger}$Imperial College London\\
\texttt{berivan.isik@stanford.edu}, \texttt{pasefrance@dei.unipd.it}
}

\iclrfinalcopy 
\begin{document}

\maketitle
\begin{abstract}
One main challenge in federated learning is the large communication cost of exchanging weight updates from clients to the server at each round. While prior work has made great progress in compressing the weight updates through gradient compression methods, we propose a radically different approach that does not update the weights at all. Instead, our method freezes the weights at their initial \emph{random} values and learns how to sparsify the random network for the best performance. To this end, the clients collaborate in training a \emph{stochastic} binary mask to find the optimal sparse random network within the original one. At the end of the training, the final model is a sparse network with random weights -- or a subnetwork inside the dense random network. We show improvements in accuracy, communication (less than $1$ bit per parameter (bpp)), convergence speed, and final model size (less than $1$ bpp) over relevant baselines on MNIST, EMNIST, CIFAR-10, and CIFAR-100 datasets, in the low bitrate regime. 

\end{abstract}
\section{Introduction}
\label{sec:introduction}

\Gls{fl} is a distributed learning framework where clients collaboratively train a model by performing local training on their data and by sharing their local updates with a server every few iterations, which in turn aggregates the local updates to create a global model, that is then transmitted to the clients for the next round of training. While being an appealing approach for enabling model training without the need to collect client data at the server, \emph{uplink} communication of local updates 
is a significant bottleneck in \gls{fl} \citep{kairouz2021advances}. This has motivated research in communication-efficient \gls{fl} strategies \citep{mcmahan2017communication} and various gradient compression schemes via sparsification \citep{lin2018deep, wang2018atomo, barnes2020rtop, ozfatura2021time, isik2022information}, quantization \citep{alistarh2017qsgd, wen2017terngrad, bernstein2018signsgd, mitchell2022optimizing}, and low-rank approximation \citep{konevcny2016federated, vargaftik2021drive, vargaftik2022eden, basat2022quick}. In this work, while aiming for communication efficiency in \gls{fl}, we take a radically different approach from prior work, and propose a strategy that does not require communication of  weight updates. To be more precise, instead of training the weights,

(1) the server initializes a dense random network with $d$ weights, denoted by the weight vector $\bm{w^{\text{init}}} =(w^{\text{init}}_1, w^{\text{init}}_{2}, \dots, w^{\text{init}}_{d})$, using a random seed $\mathsf{SEED}$, and broadcasts $\mathsf{SEED}$ to the clients enabling them to reproduce the same $\bm{w^{\text{init}}}$ locally,
 
(2) both the server and the clients keep the weights frozen at their initial values $\bm{w^{\text{init}}}$ at all times,

(3) clients collaboratively train a \emph{probability mask} of $d$ parameters $\bm{\theta} = (\theta_1, \theta_2, \dots, \theta_d) \in [0, 1]^d$, 
    
(4) the server samples a binary mask from the trained probability mask and generates a sparse network with random weights -- or a subnetwork inside the initial dense random network as follows
    \begin{equation}
    \bm{w^{\text{final}}} = \text{Bern}(\bm{\theta}) \odot  \bm{w^{\text{init}}},
\end{equation}
    where $\text{Bern}(\cdot)$ is the Bernoulli sampling operation and $\odot$ the element-wise multiplication.

\begin{figure}[h]
 \vspace{-2mm}
 \centering
\includegraphics[width=0.8\textwidth]{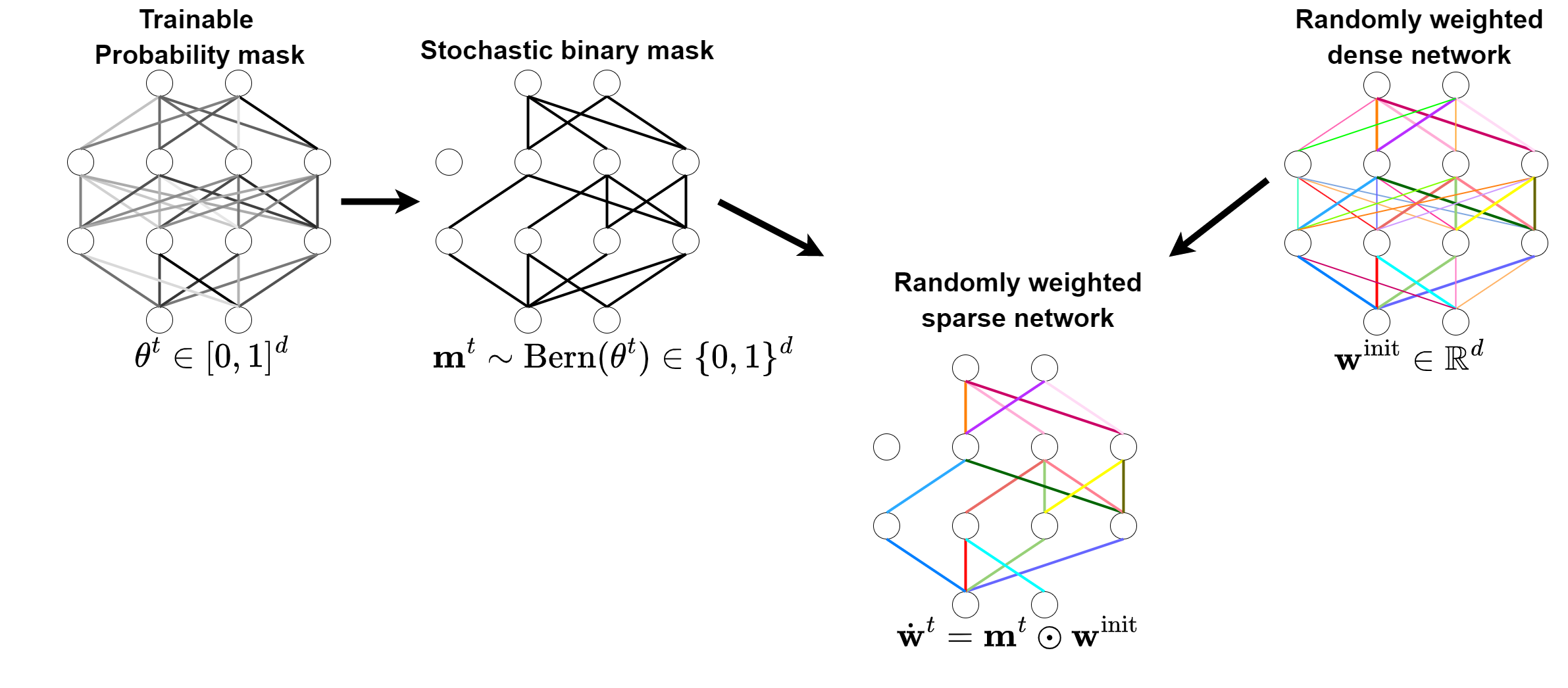}
\caption{Extracting a randomly weighted sparse network using the trainable probability mask $\bm{\theta}^{t}$ in the forward-pass of round $t$ (for clients and the server). In practice, clients collaboratively train continuous scores $\bm{s} \in \mathbb{R}^d$, and then at inference time, the clients (or the server) find $\bm{\theta}^{t} = \text{Sigmoid}(\bm{s}^{t}) \in [0,1]^d$. We skip this step in the figure for the sake of simplicity. }\label{fig:masking_diagram}
\vspace{-3.5mm}
\end{figure}

We call the proposed framework \gls{fedpm} and summarize it in Figure~\ref{fig:masking_diagram}.
At first glance, it may seem surprising that there exist subnetworks inside randomly initialized networks that could perform well without ever modifying the weight values. This phenomenon has been explored to some extent in prior work \citep{zhou2019deconstructing, ramanujan2020s, pensia2020optimal, diffenderfer2020multi, aladago2021slot} with different strategies for finding the subnetworks. 
However, how to find these subnetworks in a \gls{fl} setting has not attracted much attention so far. Some exceptions to this are works by \cite{li2021fedmask, vallapuram2022hidenseek,  mozaffari2021frl}, which provide improvements in other \gls{fl} challenges, such as personalization and poisoning attacks, while not being competitive with existing (dense) compression methods such as QSGD~\citep{alistarh2017qsgd}, DRIVE~\citep{vargaftik2021drive}, and SignSGD~\citep{bernstein2018signsgd} in terms of \emph{accuracy} under the same communication budget. In this work, we propose a \emph{stochastic} way of finding such subnetworks while reaching higher accuracy at a reduced communication cost -- less than 1 bit per parameter (bpp).

In addition to the accuracy and communication gains, our framework also provides an efficient representation of the final model post-training by requiring less than 1 bpp to represent (i) the random seed that generates the initial weights $\bm{w^{\text{init}}}$, and (ii) a sampled binary vector $\text{Bern}(\bm{\theta})$ (computed with the trained $\bm{\theta}$). Therefore, the final model enjoys a memory-efficient deployment -- a crucial feature for machine learning at power-constrained edge devices. Another advantage our framework brings is the privacy amplification under some settings, thanks to the stochastic nature of our
training strategy. 

Our contributions can be summarized as follows:

(1) We propose a \gls{fl} framework, in which the clients do not train the model weights, but instead a stochastic binary mask to be used in sparsifying the dense network with random weights. This differs from the standard training approaches in the literature. 

(2) Our framework provides efficient communication from clients to the server by requiring (less than) $1$ bpp per client while yielding faster convergence and higher accuracy than the baselines. 

(3) We propose a Bayesian aggregation strategy at the server side to better deal with partial client participation and non-IID data splits.

(4) The final model (a sparse network with random weights) can be efficiently represented with a random seed and a binary mask which requires (less than) 1 bpp -- at least $32 \times$ more efficient storage and communication of the final model with respect to standard \gls{fl} strategies.

(5) We demonstrate the efficacy of our strategy on MNIST, EMNSIT, CIFAR-10, and CIFAR-100 datasets under both IID and non-IID data splits; and show improvements in accuracy, bitrate, convergence speed, and final model size over relevant baselines, under various system configurations.


\section{Related Work}
\label{sec:related}
In this section, we briefly discuss the related work in (1) communication-efficient \gls{fl}, (2) pruning for \gls{fl}, and (3) finding subnetworks in a random network. 

\paragraph{\textbf{Communication-Efficient FL.}}  One way of improving communication efficiency in \gls{fl} is to compress the model updates using gradient compression methods like sparsification \citep{aji2017sparse, lin2018deep, wang2018atomo, barnes2020rtop,ozfatura2021time, isik2022information}, quantization \citep{alistarh2017qsgd, wen2017terngrad, suresh2017distributed, bernstein2018signsgd, mitchell2022optimizing}, and low-rank approximation \citep{ wang2018atomo, vogels2019powersgd, vargaftik2021drive, vargaftik2022eden, mohtashami2022masked, basat2022quick}; or more FL-oriented compression schemes such as \citep{konevcny2016federated, mcmahan2017communication, sattler2019robust, rothchild2020fetchsgd, reisizadeh2020fedpaq, haddadpour2020fedsketch, haddadpour2021federated}, while training a \emph{dense} network. Our framework differs from these \emph{dense} compression methods substantially due to the unconventional stochastic mask training strategy; however, we take SignSGD \citep{bernstein2018signsgd}, TernGrad \citep{wen2017terngrad}, QSGD \citep{alistarh2017qsgd}, DRIVE \citep{vargaftik2021drive}, and EDEN \citep{vargaftik2022eden} as our baselines since they work in the same bitrate regime as ours ($\approx$1 bpp). 

\paragraph{\textbf{Pruning for FL.}} Since the introduction of the \gls{lth}~\citep{frankle2018lottery}, there has been growing interest in finding sparse and \emph{trainable} networks at initialization. The main hypothesis in this line of work is that there exist sparse networks (lottery tickets) inside randomly initialized dense networks such that those sparse networks can be \emph{trained} to a surprisingly good performance. 
In the original paper, the strategy for finding these lottery tickets is to iteratively train the dense network, i.e., finding the lottery tickets is expensive. 
We distinguish our approach from the \gls{fl} papers that utilize the \gls{lth} \citep{li2020lotteryfl, ji2020dynamic, seo2021communication} and pruning \citep{lin2020esmfl, munir2021fedprune, yu2021adaptive, liu2021fedprune, jiang2022model, babakniya2022federated, dai2022dispfl, lin2020esmfl, bibikar2022federated} for three mains reasons: (i) These methods require training the weight values, and thus cannot provide an efficient representation of the final model as \gls{fedpm} does. Recall that we achieve at least $32\times$ more efficient storage and communication of the final model by representing it with just a random seed and a binary mask. (ii) Some of these works require finding the lottery tickets prior to \gls{fl} training~\citep{li2020lotteryfl}. While this could improve the communication cost during the \gls{fl} training since they communicate sparse networks, it increases the computation cost significantly due to the burden of finding lottery tickets via training. (iii) As opposed to the \gls{lth}- or pruning-based \gls{fl} works, our framework learns with what probability a particular weight should stay in the final model, i.e., the final sparsity level is also a learned parameter optimized for the best performance. Overall, since \gls{fedpm} is not in the same bitrate regime as these works (they require higher bitrates to communicate continuous weight values), we do not compare against them.

\paragraph{\textbf{Finding Subnetworks Inside a Random Network.}} Our work is closest to recent works of \cite{zhou2019deconstructing, ramanujan2020s, pensia2020optimal, aladago2021slot}, which find subnetworks (or supermasks) inside a dense network with \emph{random} weights that perform surprisingly well without ever training the weights, but in a centralized scenario. In this work, we take advantage of the existence of such subnetworks to \textbf{reduce the communication budget} in \gls{fl} to \textbf{less than 1 bpp} with \textbf{faster convergence} and \textbf{higher accuracy} than our relevant baselines in the same bitrate regime, while \textbf{further compressing the final model}, all simultaneously. Prior works \citep{li2021fedmask, vallapuram2022hidenseek, mozaffari2021frl} also consider finding subnetworks inside a dense random network in a \gls{fl} setting, but they differ from our approach on several levels. For instance, they focus on different challenges in \gls{fl}, such as personalization and poisoning attacks, which limits their ability to improve over existing compression methods in accuracy-communication bitrate tradeoff. One fundamental reason for this is their deterministic mask training strategy, which involves hard thresholding or sign operations. On the other hand, the stochasticity in \gls{fedpm} allows us to (i) enjoy a better accuracy-communication cost tradeoff, (ii) have an unbiased estimate of the true aggregate of the local masks with a provable upper bound on the error, (iii) design an improved aggregation strategy with a Bayesian approach so that the previous masks at the server are not hard replaced -- a useful strategy specifically in unbalanced non-IID splits, and (iv) gain privacy benefits via amplification in the Bernoulli sampling step. To demonstrate these benefits over deterministic schemes, we compare our method against FedMask~\citep{li2021fedmask} by adapting it slightly to mainly focus on communication efficiency, rather than personalization, and to improve its accuracy-communication efficiency performance. More specifically, we discard the initial pruning stage that was deployed for personalization. This change was necessary because (a) this paper does not study personalization, so this pruning step would put FedMask at a disadvantage in our experimental setup, and (b) the initial pruning step requires extra training at client devices, which is computationally more expensive than \gls{fedpm} and the \emph{dense} baselines.

\section{Federated Probabilistic Mask Training ($\mathtt{FedPM}$)}
\label{sec:method}
 We first describe the simpler version of the \gls{fedpm} framework in Section~\ref{sec:fedsparsenet}, which provides an \emph{unbiased} estimation of the mean of the learned probability masks at the server with \emph{bounded error}. Next, we propose a modification in our aggregation strategy by exploiting the underlying Bernoulli mechanism in Section~\ref{sec:bayesian_agg}. This helps boost the performance of \gls{fedpm} in the case of partial client participation. We then discuss the details of the distribution of the initial weights in Section~\ref{sec:distribution}, and finally describe the privacy benefits of \gls{fedpm} in Section~\ref{sec:privacy}. We use capital letters for random variables, small letters for their realization and deterministic quantities, and bold letters for vectors. Moreover, we indicate with $\bm{x}^{u, t}$ the state of the local vector $\bm{x}$ (e.g., the local mask) at client $u$ during round $t$, and with $x^{u, t}_i$ its $i$-th component. Global values are denoted with  $\bm{x}^{g, t}$ and $x^{g, t}_i$, and sets are indicated with calligraphic fonts. We denote a neural network with weight vector $\bm{p}$ as $f_{\bm{p}}$.


\subsection{$\mathtt{FedPM}$}
\label{sec:fedsparsenet}

In this section, we present the general \gls{fedpm} training pipeline. First, the server randomly initializes a neural network $f_{\bm{w}^{\text{init}}}$, parameterized by the weight vector $\bm{w}^{\text{init}} = (w^{\text{init}}_{1}, w^{\text{init}}_2, \dots, w^{\text{init}}_d) \in \mathbb{R}^d$, whose components are sampled IID according to a distribution $P_{\bm{w}}$ using a \emph{randomly generated} seed $\mathsf{SEED}$. The random $\mathsf{SEED}$ value is then communicated to all the clients, which can locally sample the same pseudo-random vector $\bm{w}^{\text{init}}$, which is kept fixed and never modified during training. The goal for the clients is to collaboratively train a probability mask $\bm{\theta} \in [0,1]^d$, which indicates the Bernoulli parameters for the global stochastic binary mask $\bm{M} \sim \text{Bern}(\bm{\theta}) \in \{0,1\}^d$, 
such that the function $f_{\bm{\dot{W}}}$ maximizes its performance on a given task, 
where $\bm{\dot{W}} = \bm{M} \odot \bm{w}^{\text{init}}$. Specifically, \gls{fedpm} learns the probabilities for the weights of being active, which are given by the probability mask $\bm{\theta} = (\theta_1, \theta_2, \dots, \theta_d) \in [0,1]^d$. To achieve this, at every round $t$, the server samples a set $\mathcal{K}_t$ of $|\mathcal{K}_t| = K $ participants (out of the total $N$ clients), which individually train their local probability masks $\bm{\theta}^{k, t}, k \in \mathcal{K}_t$, by using their local datasets $\mathcal{D}_k$, each composed of $D_k = |\mathcal{D}_k|$ samples. These local masks are then aggregated by the server in a communication-efficient way to estimate the optimal $\bm{\theta}$.
At test time, at the server, the initial random network $f_{\bm{w}^{\text{init}}}$ is sparsified using the global probability mask $\bm{\theta}^{g, t}$, following the stochastic approach in Figure~\ref{fig:masking_diagram}.
In the following sections, we provide more details on each step of each round. We give the pseudocode for \gls{fedpm} in Appendix~\ref{sec:algorithm_app}.

\subsubsection{Local Training of Probability Masks}
\label{sec:local_train}
Upon receiving a global probability mask $\bm{\theta}^{g, t-1}$ from the server at the beginning of round $t$, the client $k$ performs local training and updates the mask via back-propagation. First, however, we have to guarantee that the updated probability mask satisfies $\bm{\theta}^{k, t} \in [0,1]^d$. While this can be achieved with a regularization term in the loss, this may require clipping $\bm{\theta}^{k, t} \in [0,1]^d$ before taking a Bernoulli sample, especially in the early training stages. Clipping would then make the estimate at the server biased and hence lead to a slower convergence and lower accuracy. Therefore, similarly to the work of \cite{zhou2019deconstructing}, we introduce another mask, called \textit{score mask} $\bm{s}=(s_1, s_2, \dots, s_d ) \in \mathbb{R}^d$, that has unbounded support and can be used to generate the probability masks through the one-to-one sigmoid function by setting $\bm{\theta} = \text{Sigmoid}(\bm{s})$. Then, the procedure for local training of the probability mask at round $t$ is as follows (here, the steps from Step 2 to 4 describe one local iteration, which is repeated a number $\tau$ of times as standard in FL~\citep{mcmahan2017communication}):

\textbf{(1)} The server sends the global probability mask $\bm{\theta}^{g, t-1}$ to $K$ chosen clients, and the clients set $\bm{s}^{k, t} = \text{Sigmoid}^{-1}(\bm{\theta}^{g, t-1})$, where $\text{Sigmoid}^{-1}(\cdot)$ is the inverse of the sigmoid function.



\textbf{(2)} Then, the clients generate a binary mask by first transforming back $\bm{\theta}^{k, t} = \text{Sigmoid}(\bm{s}^{k, t})$, and then sampling a binary mask $\bm{M}^{k, t}$ from $\bm{\theta}^{k, t}$ as shown in Figure~\ref{fig:masking_diagram}: ${\bm{m}^{k, t} \sim \text{Bern}(\bm{\theta}^{k, t}).}$
    
\textbf{(3)} The sampled binary mask then  sparsifies the initial weight vector $\bm{w}^{\text{init}}$: $\dot{\bm{w}}^{k, t} = \bm{m}^{k, t} \odot \bm{w}^{\text{init}}.$

\textbf{(4)} $\dot{\bm{w}}^{k, t}$ is then used for forward pass, and the loss  $\mathcal{L}(f_{\dot{\bm{w}}^{k, t}}, \mathcal{D}_k)$ on the local task is backpropagated to update the score mask as $\bm{s}^{k, t} = \bm{s}^{k, t} - \eta \nabla \mathcal{L}(f_{\dot{\bm{w}}^{k, t}}, \mathcal{D}_k)$ ($\eta$ is the local learning rate).

All the local operations from Step 2 to Step 4 are differentiable, except for the Bernoulli sampling. We backpropagate the gradients through the Bernoulli sampling operation with a straight-through estimator \cite{bengio2013estimating}, using the first-order gradient of the Bernoulli function, which is simply equal to the probability mask $\bm{\theta}^{k, t}$. 

\subsubsection{Communication Strategy}
\label{sec:communication}
\begin{figure}[!t]
 \centering
\includegraphics[width=.6\textwidth]{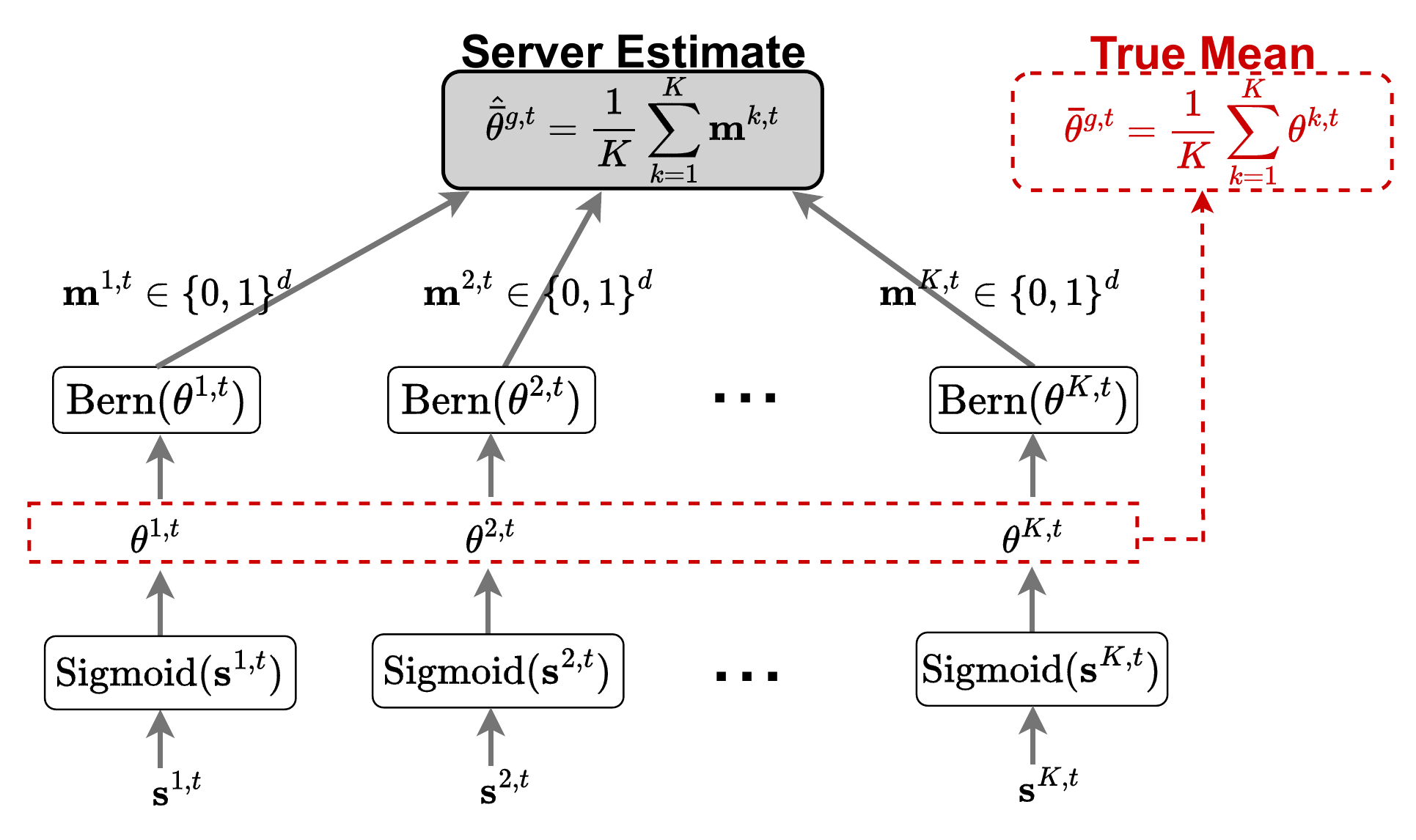}
\caption{Communication-efficient estimation of the mean of the probability masks $\bar{\bm{\theta}}^{g, t}$. Each client communicates a stochastic binary mask $\bm{m}^{k, t}$ sampled from the local probabillity mask $\bm{\theta}^{k, t}$. We reduce the bitrate to less than 1 bit per parameter by using arithmetic coding to encode $\bm{m}^{k, t}$. When the frequency of $1$'s is far from $0.5$ (which is usually the case with \gls{fedpm}), the number of bits per parameter to communicate $\bm{m}^{k, t}$ is less than 1. See Figure~\ref{fig:comparison_iid} for more details. 
}\label{fig:dist_mean_est_diagram}
\end{figure}
Once the local training at round $t$ is completed, the server needs to distill the global probability mask $\bm{\theta}^{g, t}$, by taking the empirical average of the local probability masks $\bar{\bm{\theta}}^{g, t} = \frac{1}{K} \sum_{k \in \mathcal{K}_t} \bm{\theta}^{k, t}$ collected from the clients. However, since we aim for communication efficiency, the clients do not send their local probability masks directly. Instead, they communicate a stochastic binary sample $\bm{M}^{k, t} $ from their probability masks sampled as $\bm{m}^{k, t} \sim \text{Bern}(\bm{\theta}^{k, t})$, and then the server estimates the global aggregate $\bar{\bm{\theta}}^{g, t}$ as $\hat{\bar{\bm{\theta}}}^{g, t} = \frac{1}{K} \sum_{k \in \mathcal{K}_t} \bm{m}^{k, t}$. This distributed mean estimation problem with communication constraints is summarized in Figure~\ref{fig:dist_mean_est_diagram}. Our estimator $\hat{\bar{\bm{\theta}}}^{g, t} = \frac{1}{K} \sum_{k \in \mathcal{K}_t} \bm{m}^{k, t}$ is an unbiased estimate of the true aggregate, in that 
\begin{align*}
        \E_{\bm{M}^{k,t} \sim \text{Bern}(\bm{\theta}^{k,t}) \  \forall k \in \mathcal{K}_t}[\hat{\bar{\bm{\theta}}}^{g, t}] &= \E_{\bm{M}^{k,t} \sim \text{Bern}(\bm{\theta}^{k,t}) \  \forall k \in \mathcal{K}_t}\left[\frac{1}{K} \sum_{k \in \mathcal{K}_t} \bm{M}^{k, t}\right] \\
        &= \frac{1}{K} \sum_{k \in \mathcal{K}_t} \E_{\bm{M}^{k, t} \sim \text{Bern}(\bm{\theta}^{k, t})}[ M^{k, t}] \\
    &= \frac{1}{K} \sum_{k \in \mathcal{K}_t}  \bm{\theta}^{k, t} \\
    &= \bm{\bar{\theta}}^{g, t}.
\end{align*}

Moreover, the estimation error is upper bounded as (the proof is given in Appendix~\ref{sec:error_app})
\begin{align}
\begin{aligned}
       \E_{\bm{M}^{k,t} \sim \text{Bern}(\bm{\theta}^{k,t}) \  \forall k \in \mathcal{K}_t} \big [||\hat{\bar{\bm{\theta}}}^{g, t} - \bar{\bm{\theta}}^{g, t}||_2^2 \big ]  \leq \frac{d}{4K}.  
\end{aligned}
\label{eq:error_bound}
\end{align}

Since each client communicates a stochastic binary mask $\bm{M}^{k, t}$, 1 bpp is the worst-case bitrate for \gls{fedpm}. We can further reduce the bitrate to less than 1 by using arithmetic coding \citep{rissanen1979arithmetic} or universal coding \citep{krichevsky1981performance, barron1998minimum} to encode $\bm{m}^{k, t}$, and achieve the empirical entropy since $d$ is large. This gives us smaller bitrates whenever the frequency of 1's in $\bm{m}^{k, t}$ is far from 0.5 -- which is usually the case for our method (see Figure~\ref{fig:comparison_iid} and Appendix~\ref{sec:bitrate_results} for results). 
We note that, with a deterministic mask training approach as in FedMask \citep{li2021fedmask}, arithmetic coding of $\bm{m}^{k, t}$s does not provide any further gain in bitrate, as we have empirically observed that the frequency of 1's is always around 0.5 (see Figure~\ref{fig:comparison_iid} and Appendix~\ref{sec:bitrate_results}) -- here we apply arithmetic coding for FedMask to improve our baseline although it was not proposed in the original paper. Moreover, FedMask \citep{li2021fedmask} and HideNSeek \citep{vallapuram2022hidenseek} do not enjoy the guarantees we have as their estimator (i) is not unbiased and  (ii) does not have an upper bound on the estimation error due to hard thresholding \citep{li2021fedmask} and sign operations \citep{vallapuram2022hidenseek}. This is another benefit of our stochastic sampling approach.

\subsection{$\mathtt{FedPM}$ with Bayesian Aggregation}
\label{sec:bayesian_agg}

Another important aspect that differentiates our work from existing masking methods such as FedMask~\citep{li2021fedmask} and HideNSeek~\citep{vallapuram2022hidenseek} is the Bayesian aggregation strategy, which exploits the underlying \emph{stochastic mask} to synthesize a global model, boosting the performance in scenarios where only a fraction of the clients participate in each round. Given the probabilistic interpretation of the \gls{fedpm} mask's values, at the server side we further model the probability mask $\bm{\theta}^{g, t}$ with a Beta distribution $\text{Beta}(\bm{\alpha}^{g, t}, \bm{\beta}^{g, t})$, parameterized by the round-dependent parameters $\bm{\alpha}^{g, t}$ and $\bm{\beta}^{g, t}$, which are initialized to $\bm{\alpha}^{g, 0} = \bm{\beta}^{g, 0} = \bm{\lambda_0}$. At the beginning of the training process, there is no prior knowledge indicating which network weight should be more important than the others, and so each entry in the probability mask is uniformly distributed in $[0, 1]$ -- which is the \emph{prior} distribution. Consequently, the clients' local binary masks $\bm{M}^{k, t}$s are the \emph{data} the server uses to update its belief on each weight score, and so the aggregation strategy corresponds now to a  \emph{posterior update}. Specifically, given the conjugate relation between the Beta-Bernoulli distributions, the new posteriors are still Beta distributions with parameters
\begin{align}
    \bm{\alpha}^{g, t} = \bm{\alpha}^{g, t-1} + \bm{M}^{\text{agg}, t} \quad \text{and} \quad \bm{\beta}^{g, t} = \bm{\beta}^{g, t-1} + K \cdot \bm{1} - \bm{M}^{\text{agg}, t} \quad \forall t \geq 1,
\end{align}
where $\bm{M}^{\text{agg}, t} = \sum_{k \in \mathcal{K}_t} \bm{M}^{k, t}$, and $\bm{1}$ is the $d$-dimensional all-ones vector. Then, the server broadcasts to the clients the mode of the Bernoulli distributions, as suggested by \cite{ferreira2021bayssgd}, 
\begin{align}
    \bm{\theta}^{g, t} = \frac{\bm{\alpha}^{g, t} - 1}{\bm{\alpha}^{g, t} + \bm{\beta}^{g, t} - 2},
\end{align}
where the division operation is applied element-wise. However, to obtain the best performance out of this method, the Beta parameters should be re-initialized to their original values $\bm{\lambda}_0$ with some regularity. We present an ablation study to demonstrate the improvements gained by the Bayesian aggregation strategy and the reasonable choices for the resetting frequency in Section~\ref{sec:ablation_study}. Notice that if $\bm{\lambda}_0 = \bm{1}$, and if $\bm{\alpha}$ and $\bm{\beta}$ are re-initialized at the beginning of each round, the method is equivalent to the aggregation strategy detailed in Section~\ref{sec:communication}.

\subsection{Weight Distribution}
\label{sec:distribution}
As mentioned in Section~\ref{sec:fedsparsenet}, the fixed weight vector $\bm{w}^{\text{init}}$ is initialized by sampling from the distribution $P_w$ using the randomly generated $\mathsf{SEED}$. We note that the choice of this distribution impacts two important aspects of \gls{fedpm}: (i) the values of $\bm{w}^{\text{init}}$ highly influence the final accuracy achieved by the model, as they represent the building blocks to extract a subnetwork $f_{\dot{\bm{w}}}$ (see Figure~\ref{fig:masking_diagram}), which should be rich enough to solve the learning task, and (ii) the size of the sample space of $P_w$ affects the number of bits needed to store the model during the \emph{inference process} (this is different from the 1 bpp model storage when the model is not in use). Regarding (i), as also proposed in \cite{ramanujan2020s}, we sample weights from a uniform distribution, whose domain is $\{-\sigma, +\sigma\}$, where $\sigma$ is the standard deviation of the Kaiming Normal distribution \citep{kaiming2015Init}. In this way, we control the variance of the neurons' output to be $\sim1$, which avoids the vanishing or the explosion of activation values. Previous experiments in \citep{zhou2019deconstructing, ramanujan2020s} also demonstrate the superior performance achieved by binary weights distributions when compared to standard continuous counterparts, e.g., Gaussian. Regarding (ii), even if knowing the value of $\mathsf{SEED}$ is enough to perfectly reconstruct the vector $\bm{w}^{\text{init}}$, one would have to generate the entire vector at every inference step. Consequently, to achieve fast inference, the actual values of the weights need to be stored in the memory of the devices during the \emph{inference process}. Fortunately, our initialization allows for efficient storage even during inference since (after reconstructing $\bm{w}^{\text{final}}$ using $\mathsf{SEED}$ and $\bm{m}^{\text{final}} \in \{0,1\}^d$) we only need to indicate whether the weight values in $\bm{w}^{\text{final}}$ are $-\sigma$, $0$, or $+\sigma$, with a ternary representation that can be efficiently deployed on hardware \citep{alemdar2017ternary}.



\subsection{Privacy}
\label{sec:privacy}
Privacy is another challenge in \gls{fl} as the model updates (in our case, $\bm{M}^{k, t}$s) may leak information about the client data. In Appendix~\ref{sec:privacy_app}, we analyze the differential privacy guarantees of \gls{fedpm} and give an initial foray into how \gls{fedpm} can be helpful in amplifying privacy.

\section{Experiments}
\label{sec:experiments}
In this section, we empirically show the performance of \gls{fedpm} in terms of accuracy, bitrate, converge speed, and the final model size. We consider four datasets: CIFAR-10 with 10 classes, CIFAR-100 \citep{krizhevsky2009learning} with 100 classes, MNIST \citep{deng2012mnist} with 10 classes, and EMNIST \citep{cohen2017emnist} with 47 classes. For CIFAR-100, we use a 10-layer convolutional network (CNN) $\mathtt{CONV}$-$\mathtt{10}$ and ResNet-18~\cite{he2016deep}; for CIFAR-10, a 6-layer CNN $\mathtt{CONV}$-$\mathtt{6}$ and ResNet-18~\cite{he2016deep}; and for MNIST and EMNIST, a 4-layer CNN $\mathtt{CONV}$-$\mathtt{4}$. A detailed description of the architectures can be found in Appendix~\ref{sec:additional_details_app}. Due to limited space, we provide the results on ResNet-18 in Appendix~\ref{sec:resnet_app}. We first compare \gls{fedpm} with SignSGD~\citep{bernstein2018signsgd}, TernGrad~\citep{wen2017terngrad}, QSGD~\citep{alistarh2017qsgd}, DRIVE~\citep{vargaftik2021drive}, EDEN~\citep{vargaftik2022eden}, and FedMask~\citep{li2021fedmask} on IID data split and full client participation in Section~\ref{sec:iid_exp}. We then extend our experiments to non-IID data splits and partial participation in Section~\ref{sec:noniid_exp}. Finally, in Section~\ref{sec:ablation_study}, we present a key ablation study to justify why the Bayesian aggregation strategy is necessary for partial participation and to demonstrate how the resetting frequency affects the convergence rate and the final accuracy. Clients perform 3 local epochs in all experiments. We provide additional details on the experimental setup in Appendix~\ref{sec:additional_details_app}. We present results averaged over 3 runs.

\subsection{IID Data Split and Full Participation ($K=N$)}
\label{sec:iid_exp}
\begin{figure}[!h]
 \centering
\includegraphics[width=.231\textwidth]{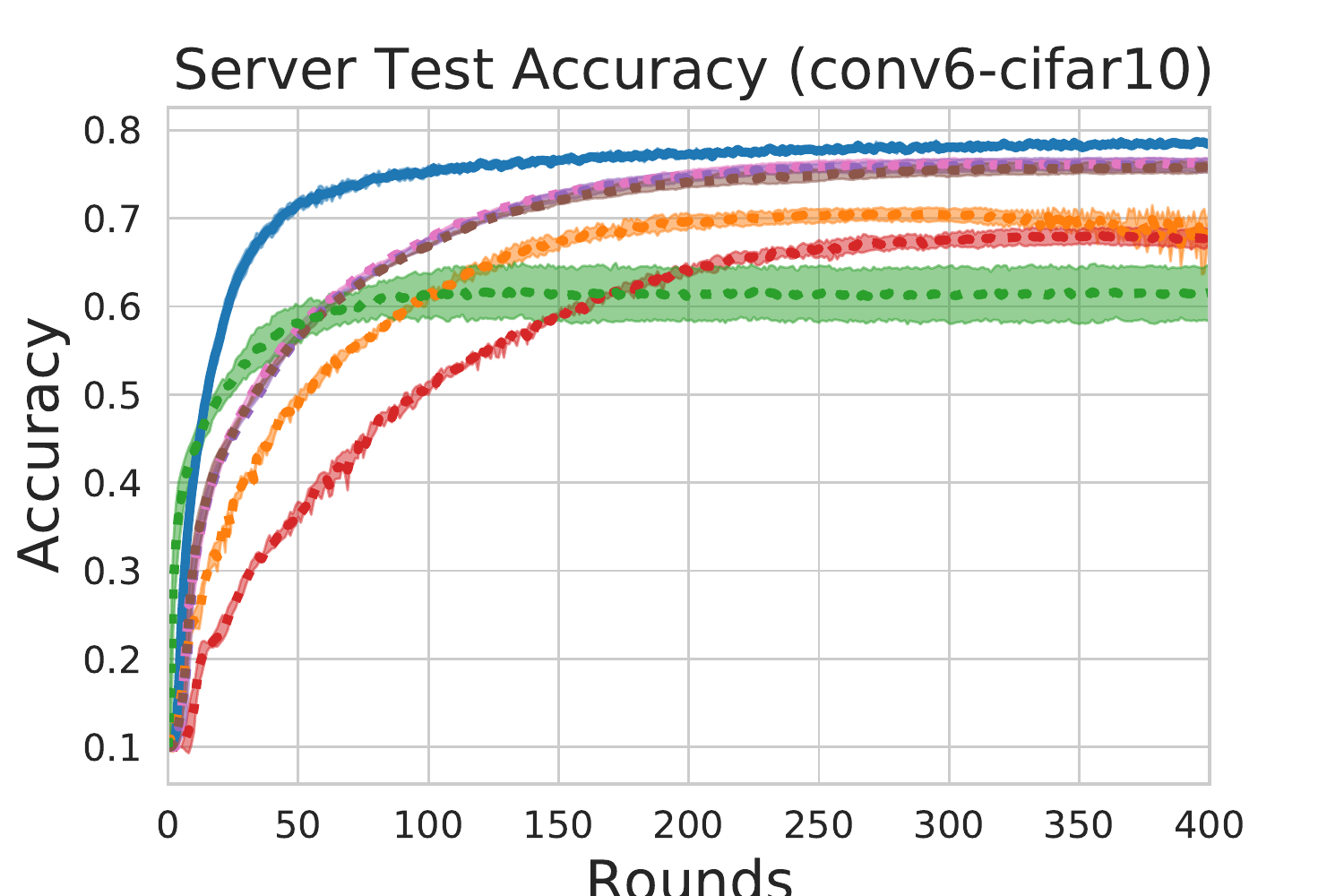}
\includegraphics[width=.226\textwidth]{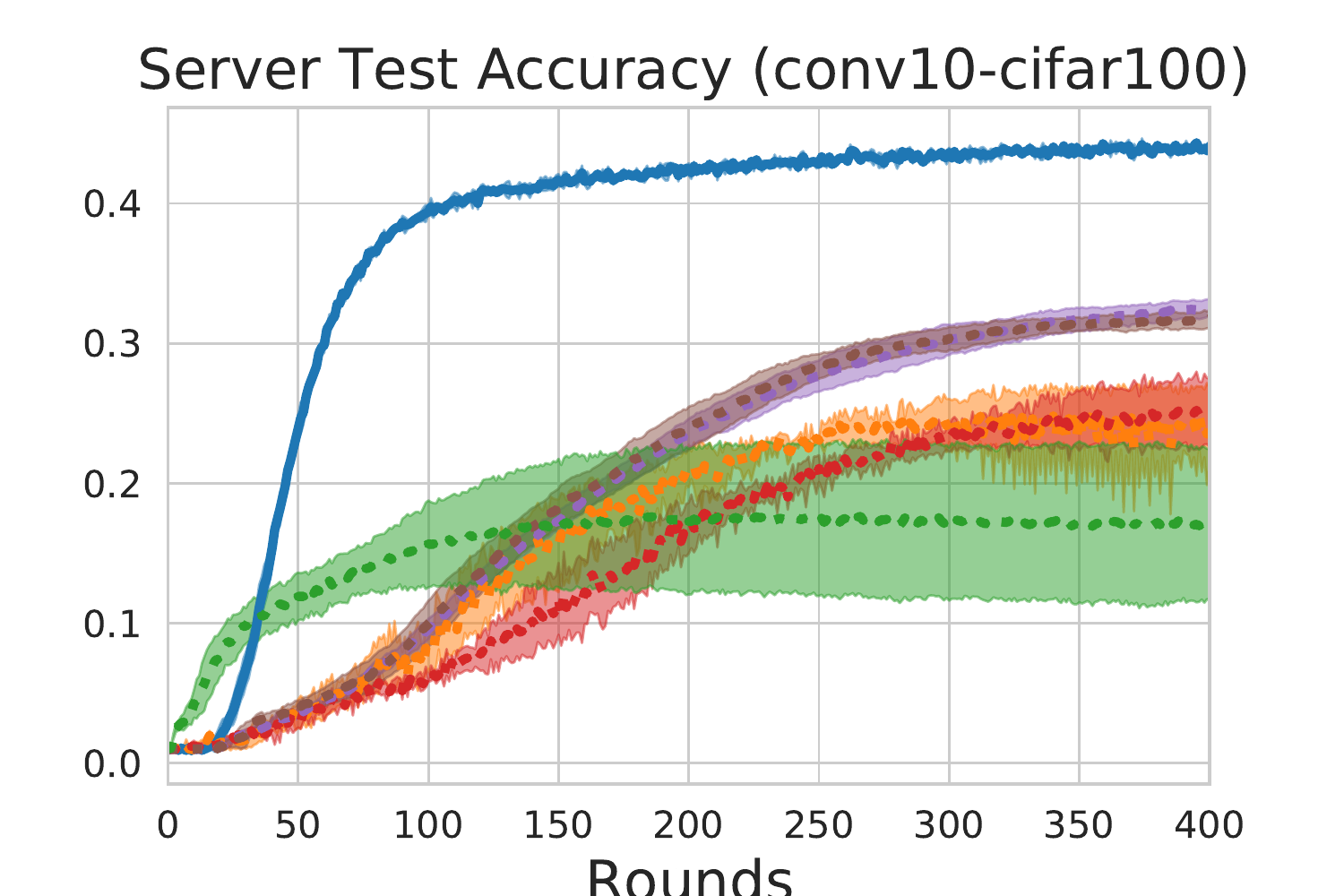}
\includegraphics[width=.226\textwidth]{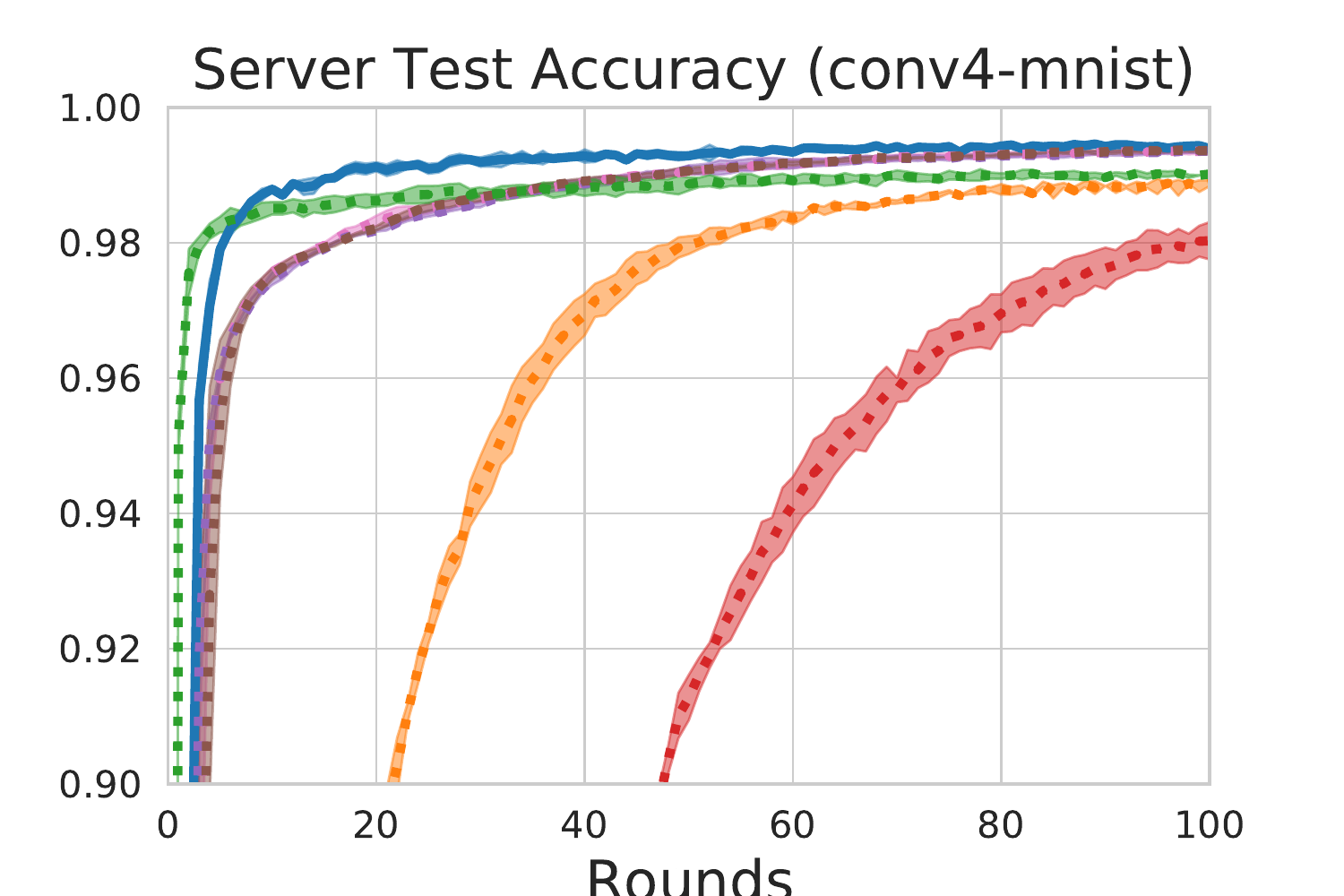}
\includegraphics[width=.29\textwidth]{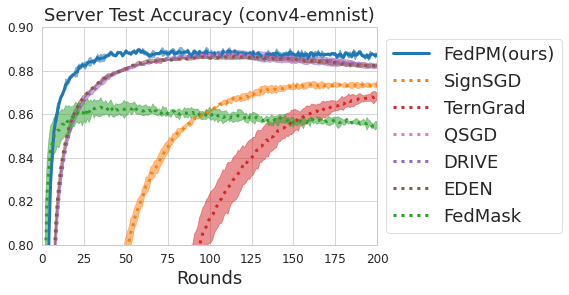}
\includegraphics[width=.231\textwidth]{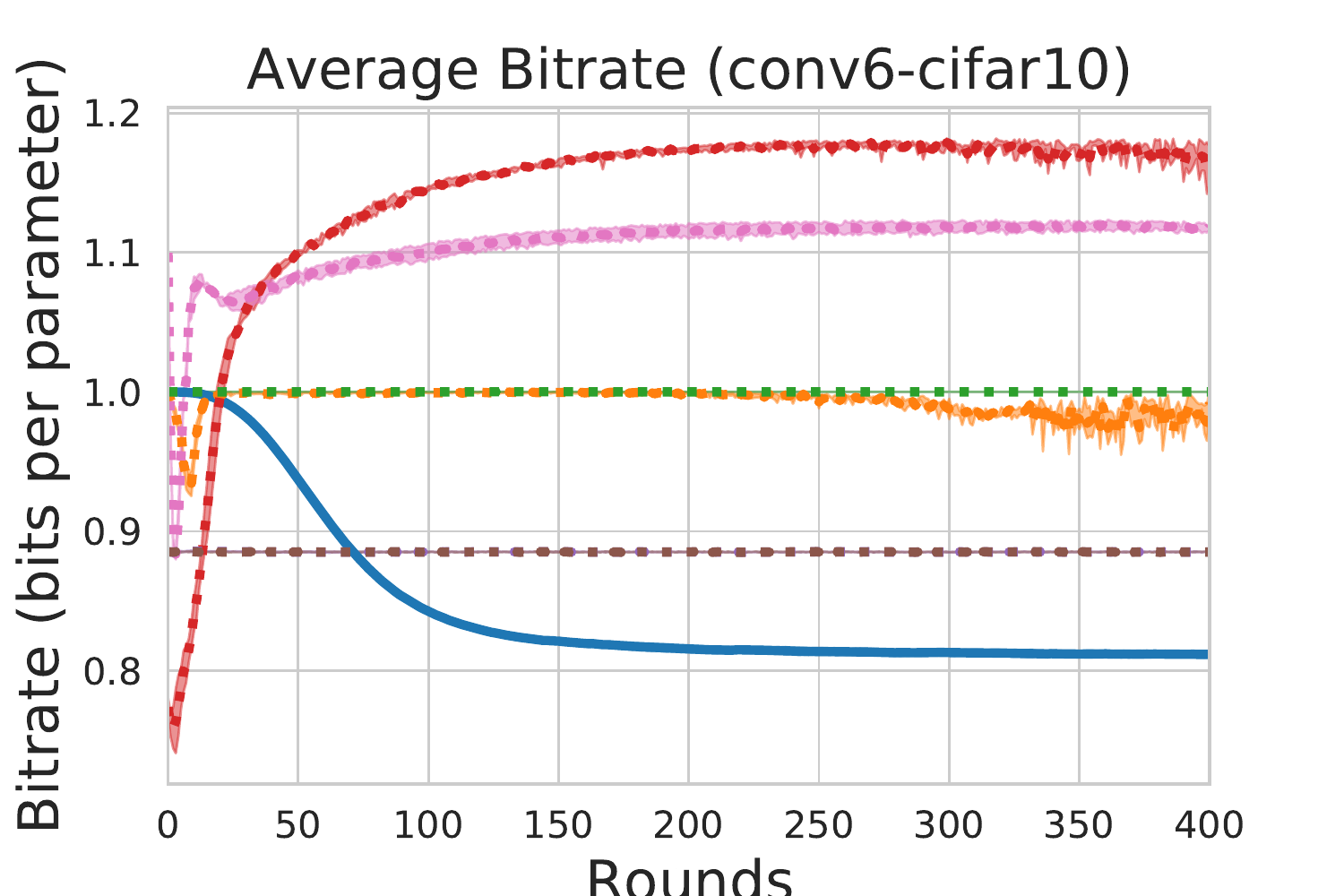}
\includegraphics[width=.226\textwidth]{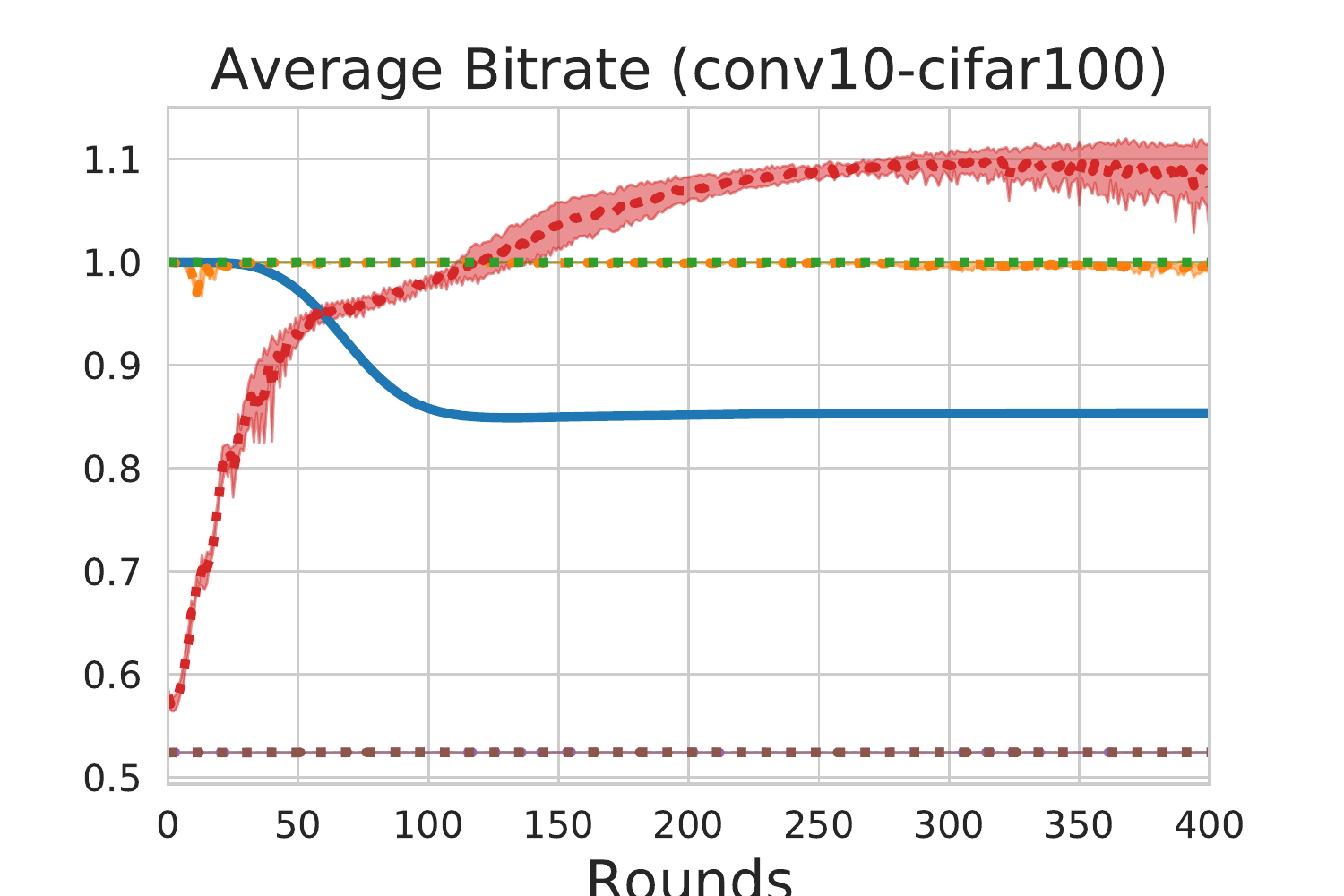}
\includegraphics[width=.226\textwidth]{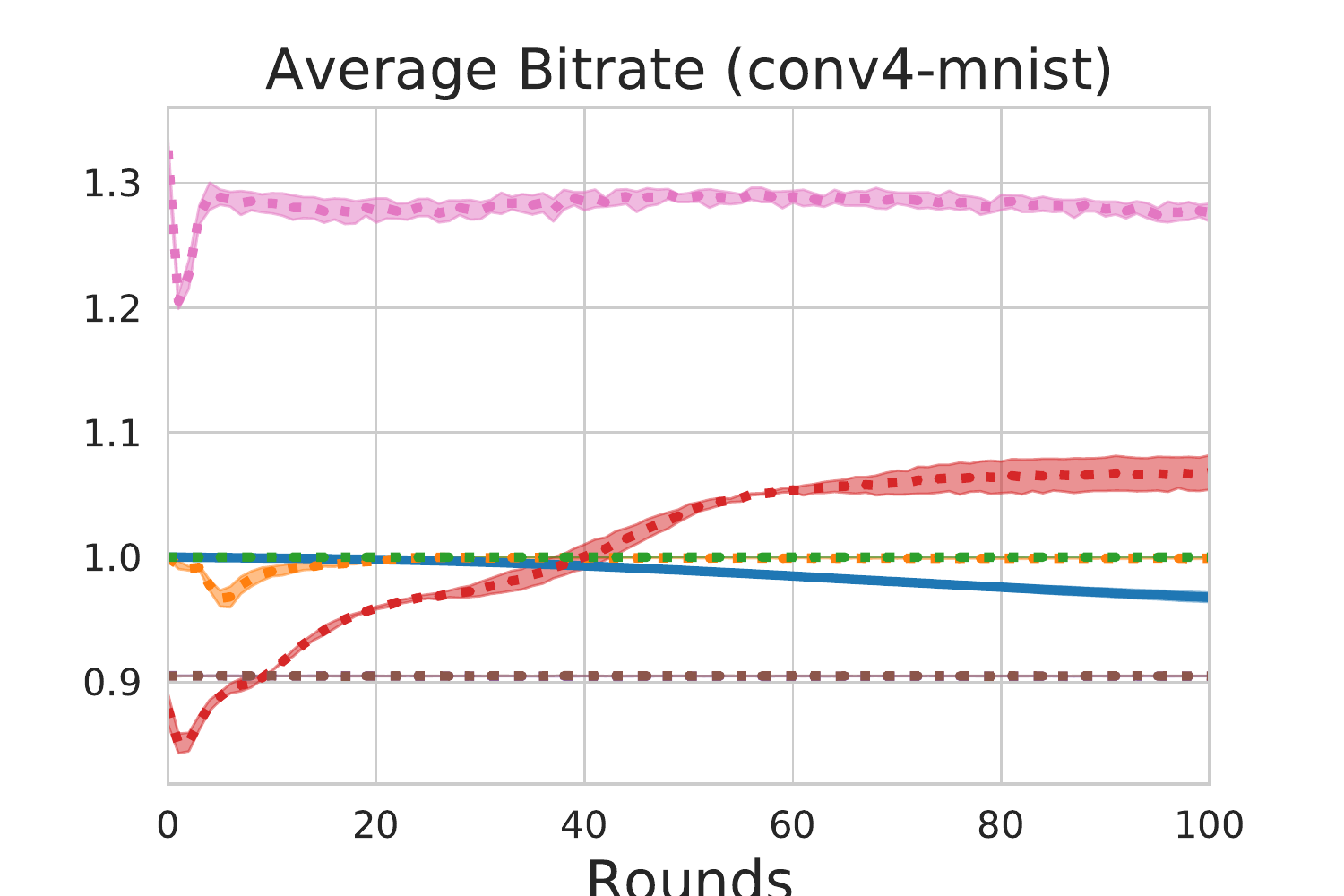}
\includegraphics[width=.29\textwidth]{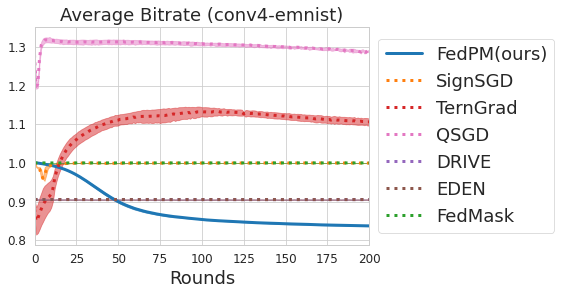}
\caption{ Accuracy and bitrate comparison of \gls{fedpm} with SignSGD~\citep{bernstein2018signsgd}, TernGrad~\citep{wen2017terngrad}, QSGD~\citep{alistarh2017qsgd}, DRIVE~\citep{vargaftik2021drive}, EDEN~\citep{vargaftik2022eden}, and FedMask~\citep{li2021fedmask}, all performing in the same  bitrate regime. 
}\label{fig:comparison_iid}
\end{figure}

In this section, we focus on IID data distribution and the case when all the clients participate in the training at each round. We set the number of clients to $N=K=10$. We report the estimated bitrate for the arithmetic code that uses the empirical frequency of the symbols (for our method \gls{fedpm}, this corresponds to the frequency of 1's in $\bm{m}^{k,t}$) -- which is equal to the empirical entropy for blocklength $d$ as large as the model size. 
In Figure~\ref{fig:comparison_iid}, we compare the accuracy, bitrate, and convergence speed of \gls{fedpm} with relevant baselines. As can be seen in the figure, \gls{fedpm} converges to the highest accuracy on all four datasets. DRIVE, EDEN, and QSGD (they mostly overlap in the accuracy plots) seem to be the three baselines that perform the best after \gls{fedpm}; however, their convergence speed is significantly lower than \gls{fedpm}. In terms of convergence speed, FedMask is the fastest among the baselines -- in fact, at the beginning of the training, FedMask is faster than \gls{fedpm} as well. However, its final accuracy is lower than the others. We also would like to highlight that while some of our baselines, such as FedMask and TernGrad, have a visibly high variance in accuracy, \gls{fedpm} shows stable training behavior across all experiments.

In terms of bitrate, SignSGD and FedMask consistently spend 1 bpp, which is the default number when a binary mask or sign mask is communicated. This means binary values (1's and 0's) are almost equally distributed in their masks, which prevents them from enjoying additional bitrate gains. Across all experiments, TernGrad has the highest bitrate. We would like to leave a note about the bitrate of QSGD. Unlike other baselines, including our work, QSGD can go down to very low bitrates by adjusting the number of levels in quantization. We have observed that in the extreme quantization case, QSGD underperforms \gls{fedpm}. Then, we have decided to increase the number of quantization levels in QSGD to see if it improves the accuracy. However, as can be seen from the plots, even with bitrate larger than 1, QSGD still underperforms \gls{fedpm}. The only two baselines that challenge \gls{fedpm} in terms of bitrate are DRIVE and EDEN. While \gls{fedpm} has lower bitrates on CIFAR-10 and EMNIST; DRIVE and EDEN have better bitrates on CIFAR-100 and MNIST. However, the accuracy of DRIVE and EDEN on these datasets (specifically CIFAR-100) is significantly lower than that of \gls{fedpm}, with slower convergence.

As for the final model size, \gls{fedpm} needs only 0.8 bpp for the $\mathtt{CONV}$-$\mathtt{6}$ model trained on CIFAR-10, 0.85 bpp for the $\mathtt{CONV}$-$\mathtt{10}$ model trained on CIFAR-100, 0.96 bpp for the $\mathtt{CONV}$-$\mathtt{4}$ model trained on MNIST, and 0.83 bpp for the $\mathtt{CONV}$-$\mathtt{4}$ model trained on EMNIST. On the other hand, other baselines that train a dense model, namely SignSGD, TernGrad, QSGD, DRIVE, and EDEN, would need to represent each weight with their full precision value, i.e., 32 bpp. This implies that \gls{fedpm} provides around $38.6 \times$ improvement in the storage or the communication of the final model. Since FedMask also trains a sparse model, it enjoys a similar gain in the final model size requiring 1 bpp across all the models. Due to the stochastic masking procedure and uneven distribution of $1$'s and $0$'s in the binary masks, \gls{fedpm} has up to 0.17 bpp improvement over the deterministic procedure in FedMask, which adds up to a large gain due to the huge model size.

We provide additional experimental results with ResNet-18 model on CIFAR-10 and CIFAR-100 datasets in Appendix~\ref{sec:resnet_app}; and observe similar improvements over the baselines.

\subsection{Non-IID Data Split and Partial Participation ($K<N$)}
\label{sec:noniid_exp}

\begin{table}[b!]
\vspace{- 0.2cm}
\centering
\caption{Average final accuracy $\pm \sigma$ in non-IID data split with $c_{\text{max}} = 4$~and~$2$, and client participation ratios $\rho = \{0.1, 0.2, 0.5, 1\}$, for \gls{fedpm}, FedMask, and the strongest baselines in the IID experiments: EDEN, DRIVE, and QSGD. The training duration was set to $t_{\text{max}} = 200$ rounds.}\label{tab:non_iid_4_class}
\resizebox{.80\columnwidth}{!}{
\begin{tabular}{ lccccc }
 \toprule
 & \textbf{Algorithm} & $\bm{\rho = 1}$& $\bm{\rho = 0.5}$ & $\bm{\rho = 0.2}$ & $\bm{\rho = 0.1}$ \\
 \midrule
& DRIVE~\citep{vargaftik2021drive} & $0.739 \pm 0.005$ & $0.632 \pm 0.010$ & $0.563 \pm 0.005$ & $0.405 \pm 0.018$ \\
& EDEN~\citep{vargaftik2022eden} & $0.717 \pm 0.006$ & $0.665 \pm 0.012$ & $0.565 \pm 0.009$ & $0.360 \pm 0.016$ \\
$c_{\text{max}} = 4$&  QSGD~\citep{alistarh2017qsgd} & $0.709 \pm 0.006$& $0.644 \pm 0.014$ & $0.567 \pm 0.010$ & $0.399 \pm 0.020$ \\
& FedMask~\citep{li2021fedmask} & $ 0.531 \pm  0.044 $ & $ 0.435 \pm  0.057$ & $ 0.434 \pm 0.036$ & $ 0.362 \pm 0.024$ \\
& \gls{fedpm} (Ours) & $\bm{0.748 \pm 0.003}$ & $\bm{0.720 \pm 0.007}$ & $\bm{0.617 \pm 0.021}$ & $\bm{0.496 \pm 0.007}$ \\
 \midrule
& DRIVE~\citep{vargaftik2021drive} & $0.434 \pm 0.025$ & $0.376 \pm 0.014$ & $\bm{0.375 \pm 0.015}$ & $0.221 \pm 0.003$ \\
& EDEN~\citep{vargaftik2022eden} & $0.535 \pm 0.050$ & $0.461 \pm 0.016$ & $\bm{0.380 \pm 0.015}$ & $0.219 \pm 0.005$ \\
$c_{\text{max}} = 2$ & QSGD~\citep{alistarh2017qsgd} &  $0.476 \pm 0.033$ & $0.464 \pm 0.002$ & $\bm{0.375 \pm 0.026}$ & $0.243 \pm 0.014$  \\
& FedMask~\citep{li2021fedmask}& $0.420 \pm 0.028 $ & $ 0.387 \pm 0.062 $ & $ 0.285 \pm 0.040$ & $ 0.197 \pm 0.030$ \\
&\gls{fedpm} (Ours) & $\bm{0.643 \pm 0.016}$ & $\bm{0.556 \pm 0.031}$ & $\bm{0.372 \pm 0.004}$ & $\bm{0.277 \pm 0.003}$ \\
 \bottomrule
\end{tabular}}
\end{table}

This section considers more realistic scenarios, in which the local clients' datasets are generated from slightly different data distributions. We focus on CIFAR-10 with $\mathtt{CONV}$-$\mathtt{6}$, and we compare \gls{fedpm} against (i) the most promising baselines, which, based on the results of Section~\ref{sec:iid_exp}, are DRIVE, EDEN, and QSGD, and (ii) FedMask, as it is the only \emph{sparse} baseline.
To choose the size of each dataset $|\mathcal{D}_n| = D_n$, for each client $n \in \{1, \dots, N\}$, an integer $j_n$ is sampled uniformly from $\{10, 11, \dots, 100\}$. Then, a coefficient $p_n = \frac{j_n}{\sum_n j_j}$ is computed, which represents the size of the local dataset $D_n$ as a fraction of the size of the full dataset, i.e., the training set of CIFAR-10. In this way, highly unbalanced datasets can be generated from the central one. Moreover, since the task is a classification problem, we impose a maximum number of different labels, or classes, $c_{\text{max}}$, that one client can see. Consequently, clients need cooperation to learn the statistics of other classes' distributions, as the test dataset contains samples from all classes. In addition, partial participation is also considered, meaning that at each round, the server uniformly samples a fraction $\rho = \frac{K}{N}$ of the clients to participate in the training round. This is motivated in real-world scenarios by the scarcity of physical communication network resources, which may limit the availability of part of the clients during one round. The maximum number of classes per local dataset is set to $c_{\text{max}} \in \{2, 4\}$, and the participation ratio is set to $\rho \in \{0.1, 0.2, 0.5, 1\}$. For $\rho = 1$ and $\rho = 0.5$, the total number of clients is set to $N=10$ (and so $K$ is equal to $10$ and $5$, respectively). For $\rho = 0.2$, we set $N=100$ (and so $K=20$), and for $\rho = 0.1$, we set $N=50$ (and so $K=5$), which is the worst scenario among all combinations, given the small amount of information the server can collect at the end of each round. When $\rho = 1$, for the \gls{fedpm} algorithm, we keep the same aggregation strategy exposed in Section~\ref{sec:communication} and Figure~\ref{fig:dist_mean_est_diagram}; and we switch to the Bayesian aggregation method (see Section~\ref{sec:bayesian_agg}) when there is partial participation, i.e., when $\rho < 1$. Indeed, applying the Bayesian aggregation method is revealed to be crucial for achieving good accuracy when $\rho < 1$ and data are non-IID, obtaining a large gain with respect to the simpler version in Section~\ref{sec:communication}, which resets the Beta priors at each round (or takes the average of the samples, as explained in Section~\ref{sec:bayesian_agg}). We elaborate more on this observation with an ablation study in Section~\ref{sec:ablation_study}. We adopt a simple heuristic schedule to reset the priors: Reset every $3$ rounds when $\rho = 0.5$ and $\rho = 0.2$, and every $10$ rounds when $\rho = 0.1$. As expected, the smaller the ratio $\rho$, the larger the number of rounds we should wait before resetting the priors to collect more information from a much more diverse pool of clients (see Section~\ref{sec:ablation_study} for a rule of thumb on the resetting frequency).

Table~\ref{tab:non_iid_4_class} reports the results with $c_{\text{max}} = 4$~and~$2$. \gls{fedpm} seems to outperform all the baselines in every configuration, as the Bayesian aggregation allows the server to collect more data before resetting the priors, which is important when clients' data distributions are non-IID, and only a fraction of the clients participate in each round. This strategy can be seen as the \gls{fedpm} counterpart of decreasing the learning rate (which we applied in the other \emph{dense} compression-based baselines, like DRIVE, EDEN, and QSGD). It is seen from Table~\ref{tab:non_iid_4_class} that FedMask~\citep{li2021fedmask} is struggling in the non-IID case, as applying a hard threshold on the scores to binarize the mask does not provide a proper way to implement multiple-rounds aggregation, emphasizing the benefit of the stochastic process in \gls{fedpm}. 
It is interesting to notice that, especially when $c_{\text{max}} = 4$, the lower the value of $\rho$, the larger the gap between \gls{fedpm} and the baselines, corroborating the fact that the Bayesian strategy can better deal with partial participation. Analysis of the communication bitrate is provided in Appendix.~\ref{sec:bitrate_results}.

\subsection{Ablation Study on the Bayesian Aggregation Strategy}
\label{sec:ablation_study}

\begin{figure}[!h]
 \centering
\includegraphics[width=.29\textwidth]{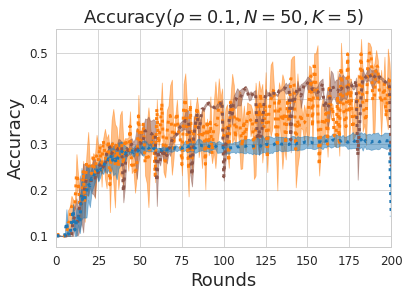}
\includegraphics[width=.27\textwidth]{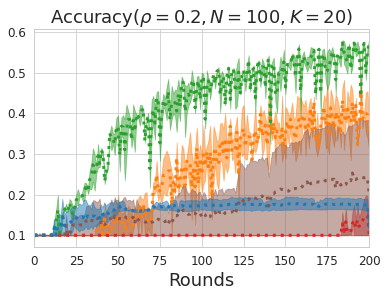}
\includegraphics[width=.36\textwidth]{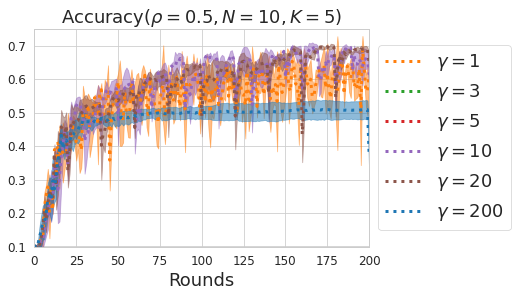}
\caption{Accuracy for different values of $\gamma$ -- the number of rounds before resetting the priors. 
}
\label{fig:ablation_study}
\vspace{- 0.5cm}
\end{figure}

In this section, we try to answer two questions: (1) \emph{Is Bayesian aggregation really necessary?} and (2) \emph{What is the effect of resetting frequency on the convergence rate and the final accuracy?} We do this by analyzing the effect of different resetting frequencies of the Beta priors on the training behavior of \gls{fedpm} with non-IID data split and partial client participation; and report the results in Figure~\ref{fig:ablation_study}. Hereafter, we denote with $\gamma$ the number of aggregation rounds before resetting the priors. For instance, $\gamma=1$ corresponds to resetting the priors at every iteration, which is equivalent to the aggregation method presented in Section~\ref{sec:communication}. On the other extreme, $\gamma=200$ indicates that the priors are never reset. It is seen that $\gamma=1$ curves fluctuate significantly and never converge to the best accuracy in any setting, while $\gamma=200$ curves look smoother but converge to the lowest accuracy in all settings. This intuitively makes sense because, as already mentioned in Section~\ref{sec:bayesian_agg}, by increasing the value of $\gamma$, we allow the server to consider the information coming from multiple rounds while updating the global parameters. Indeed, with partial participation and non-IID data, a single round's updates may convey skewed information, depending on the level of data heterogeneity $c_{\text{max}}$, and client participation ratio $\rho$. As a rule of thumb for the resetting frequency value, we suggest tuning $\gamma$ around the value $\frac{1}{\rho}$. The rationale behind this is that with uniform client sampling, at least $\frac{1}{\rho}$ rounds are needed to have the non-zero probability to sample from each client once before resetting the prior. In practice, we do not need to sample exactly from every client, as enough information is contained in the updates of the other sampled ones. 

\section{Conclusion}
\label{sec:conclusion}

In this work, we introduced Federated Probabilistic Mask Training ($\mathtt{FedPM}$) -- a communication-efficient FL strategy. $\mathtt{FedPM}$ relies on the idea of finding a sparse network in a randomly initialized dense network, which is then sparsified by a collaboratively trained \emph{stochastic} binary mask. In addition to reducing the communication cost to less than $1$ bit per parameter (bpp), $\mathtt{FedPM}$ also reaches higher accuracy with faster convergence than the relevant baselines, and can potentially amplify privacy while additionally outputting a compressed final model with a size less than 1 bpp. Throughout the manuscript, we highlighted the advantages of having a stochastic mask training approach rather than a deterministic one in terms of accuracy, bitrate, and privacy. 

\section{Ethics Statement}
All the experiments in the paper were performed on publicly available datasets. When we evaluated our strategy, we only considered accuracy as a
measure of performance. However, as pointed out by \cite{hooker2020characterising}, compression methods may disproportionately impact different subgroups of the data. We agree that this may potentially create a fairness issue in all communication-efficient federated learning frameworks and deserves more attention from the community.

\section{Reproduction Statement}
The codebase for this work is open-sourced at \url{https://github.com/BerivanIsik/sparse-random-networks}. All the hyperparameters necessary to reproduce the results in the paper can be found in Appendix~\ref{sec:additional_details_app}. We only used publicly available standard datasets and included links to them in the manuscript.
\section{Acknowledgement}
\label{sec:acknowledgement}

The authors would like to thank the anonymous reviewers and area chairs who provided valuable feedback; and Zachary Charles, Mahdi Haghifam, Peter Kairouz, and Nicole Mitchell for inspiring discussions. This work was supported in part by a Sony Stanford
Graduate Fellowship, a National Science Foundation
(NSF) award, a Meta research grant, and the European Union under the Italian National Recovery and Resilience Plan (NRRP) of NextGenerationEU, partnership on ``Telecommunications of the Future'' (PE0000001 - program ``RESTART'').

\bibliography{iclr2023_conference}

\begin{thebibliography}{70}
\providecommand{\natexlab}[1]{#1}
\providecommand{\url}[1]{\texttt{#1}}
\expandafter\ifx\csname urlstyle\endcsname\relax
  \providecommand{\doi}[1]{doi: #1}\else
  \providecommand{\doi}{doi: \begingroup \urlstyle{rm}\Url}\fi

\bibitem[Abadi et~al.(2016)Abadi, Chu, Goodfellow, McMahan, Mironov, Talwar,
  and Zhang]{abadi2016deep}
Martin Abadi, Andy Chu, Ian Goodfellow, H~Brendan McMahan, Ilya Mironov, Kunal
  Talwar, and Li~Zhang.
\newblock Deep learning with differential privacy.
\newblock In \emph{Proceedings of the ACM SIGSAC conference on computer and
  communications security}, pp.\  308--318, 2016.

\bibitem[Agarwal et~al.(2021)Agarwal, Kairouz, and Liu]{agarwal2021skellam}
Naman Agarwal, Peter Kairouz, and Ziyu Liu.
\newblock The skellam mechanism for differentially private federated learning.
\newblock \emph{Advances in Neural Information Processing Systems},
  34:\penalty0 5052--5064, 2021.

\bibitem[Aji \& Heafield(2017)Aji and Heafield]{aji2017sparse}
Alham Aji and Kenneth Heafield.
\newblock Sparse communication for distributed gradient descent.
\newblock In \emph{EMNLP 2017: Conference on Empirical Methods in Natural
  Language Processing}, pp.\  440--445. Association for Computational
  Linguistics (ACL), 2017.

\bibitem[Aladago \& Torresani(2021)Aladago and Torresani]{aladago2021slot}
Maxwell~M Aladago and Lorenzo Torresani.
\newblock Slot machines: Discovering winning combinations of random weights in
  neural networks.
\newblock In \emph{International Conference on Machine Learning}, pp.\
  163--174. PMLR, 2021.

\bibitem[Alemdar et~al.(2017)Alemdar, Leroy, Prost-Boucle, and
  P{\'e}trot]{alemdar2017ternary}
Hande Alemdar, Vincent Leroy, Adrien Prost-Boucle, and Fr{\'e}d{\'e}ric
  P{\'e}trot.
\newblock Ternary neural networks for resource-efficient {AI} applications.
\newblock In \emph{International Joint conference on Neural Networks (IJCNN)},
  pp.\  2547--2554, 2017.

\bibitem[Alistarh et~al.(2017)Alistarh, Grubic, Li, Tomioka, and
  Vojnovic]{alistarh2017qsgd}
Dan Alistarh, Demjan Grubic, Jerry Li, Ryota Tomioka, and Milan Vojnovic.
\newblock {QSGD}: Communication-efficient {SGD} via gradient quantization and
  encoding.
\newblock \emph{Advances in Neural Information Processing Systems}, 30, 2017.

\bibitem[Andrew et~al.(2021)Andrew, Thakkar, McMahan, and
  Ramaswamy]{andrew2021differentially}
Galen Andrew, Om~Thakkar, Brendan McMahan, and Swaroop Ramaswamy.
\newblock Differentially private learning with adaptive clipping.
\newblock \emph{Advances in Neural Information Processing Systems},
  34:\penalty0 17455--17466, 2021.

\bibitem[Babakniya et~al.(2022)Babakniya, Kundu, Prakash, Niu, and
  Avestimehr]{babakniya2022federated}
Sara Babakniya, Souvik Kundu, Saurav Prakash, Yue Niu, and Salman Avestimehr.
\newblock Federated sparse training: Lottery aware model compression for
  resource constrained edge.
\newblock \emph{arXiv preprint arXiv:2208.13092}, 2022.

\bibitem[Balle et~al.(2018)Balle, Barthe, and Gaboardi]{balle2018privacy}
B.~Balle, G~Barthe, and M.~Gaboardi.
\newblock Privacy amplification by subsampling: tight analyses via couplings
  and divergences.
\newblock \emph{Advances in neural information processing systems}, 2018.

\bibitem[Balle et~al.(2020)Balle, Kairouz, McMahan, Thakkar, and
  Guha~Thakurta]{balle2020privacy}
Borja Balle, Peter Kairouz, Brendan McMahan, Om~Thakkar, and Abhradeep
  Guha~Thakurta.
\newblock Privacy amplification via random check-ins.
\newblock In \emph{Advances in Neural Information Processing Systems}, 2020.

\bibitem[Barnes et~al.(2020)Barnes, Inan, Isik, and
  {\"O}zg{\"u}r]{barnes2020rtop}
Leighton~Pate Barnes, Huseyin~A Inan, Berivan Isik, and Ayfer {\"O}zg{\"u}r.
\newblock rtop-k: A statistical estimation approach to distributed {SGD}.
\newblock \emph{IEEE Journal on Selected Areas in Information Theory},
  1\penalty0 (3):\penalty0 897--907, 2020.

\bibitem[Barron et~al.(1998)Barron, Rissanen, and Yu]{barron1998minimum}
Andrew Barron, Jorma Rissanen, and Bin Yu.
\newblock The minimum description length principle in coding and modeling.
\newblock \emph{IEEE transactions on information theory}, 44\penalty0
  (6):\penalty0 2743--2760, 1998.

\bibitem[Basat et~al.(2022)Basat, Vargaftik, Portnoy, Einziger, Ben-Itzhak, and
  Mitzenmacher]{basat2022quick}
Ran~Ben Basat, Shay Vargaftik, Amit Portnoy, Gil Einziger, Yaniv Ben-Itzhak,
  and Michael Mitzenmacher.
\newblock Quick-fl: Quick unbiased compression for federated learning.
\newblock \emph{arXiv preprint arXiv:2205.13341}, 2022.

\bibitem[Bengio et~al.(2013)Bengio, L{\'e}onard, and
  Courville]{bengio2013estimating}
Yoshua Bengio, Nicholas L{\'e}onard, and Aaron Courville.
\newblock Estimating or propagating gradients through stochastic neurons for
  conditional computation.
\newblock \emph{arXiv preprint arXiv:1308.3432}, 2013.

\bibitem[Bernstein et~al.(2018)Bernstein, Wang, Azizzadenesheli, and
  Anandkumar]{bernstein2018signsgd}
Jeremy Bernstein, Yu-Xiang Wang, Kamyar Azizzadenesheli, and Animashree
  Anandkumar.
\newblock signsgd: Compressed optimisation for non-convex problems.
\newblock In \emph{International Conference on Machine Learning}, pp.\
  560--569. PMLR, 2018.

\bibitem[Bibikar et~al.(2022)Bibikar, Vikalo, Wang, and
  Chen]{bibikar2022federated}
Sameer Bibikar, Haris Vikalo, Zhangyang Wang, and Xiaohan Chen.
\newblock Federated dynamic sparse training: Computing less, communicating
  less, yet learning better.
\newblock In \emph{Proceedings of the AAAI Conference on Artificial
  Intelligence}, volume~36, pp.\  6080--6088, 2022.

\bibitem[Cohen et~al.(2017)Cohen, Afshar, Tapson, and
  Van~Schaik]{cohen2017emnist}
Gregory Cohen, Saeed Afshar, Jonathan Tapson, and Andre Van~Schaik.
\newblock {EMNIST}: Extending {MNIST} to handwritten letters.
\newblock In \emph{International Joint Conference on Neural Networks (IJCNN)},
  pp.\  2921--2926, 2017.

\bibitem[Dai et~al.(2022)Dai, Shen, He, Tian, and Tao]{dai2022dispfl}
Rong Dai, Li~Shen, Fengxiang He, Xinmei Tian, and Dacheng Tao.
\newblock Dispfl: Towards communication-efficient personalized federated
  learning via decentralized sparse training.
\newblock \emph{arXiv preprint arXiv:2206.00187}, 2022.

\bibitem[Deng(2012)]{deng2012mnist}
Li~Deng.
\newblock The {MNIST} database of handwritten digit images for machine learning
  research.
\newblock \emph{IEEE Signal Processing Magazine}, 29\penalty0 (6):\penalty0
  141--142, 2012.

\bibitem[Diffenderfer \& Kailkhura(2020)Diffenderfer and
  Kailkhura]{diffenderfer2020multi}
James Diffenderfer and Bhavya Kailkhura.
\newblock Multi-prize lottery ticket hypothesis: Finding accurate binary neural
  networks by pruning a randomly weighted network.
\newblock In \emph{International Conference on Learning Representations}, 2020.

\bibitem[Dwork et~al.(2006)Dwork, McSherry, Nissim, and
  Smith]{dwork2006calibrating}
Cynthia Dwork, Frank McSherry, Kobbi Nissim, and Adam Smith.
\newblock Calibrating noise to sensitivity in private data analysis.
\newblock In \emph{Theory of cryptography conference}, pp.\  265--284.
  Springer, 2006.

\bibitem[Erlingsson et~al.(2019)Erlingsson, Feldman, Mironov, Raghunathan,
  Talwar, and Thakurta]{Erlingsson2019amplification}
\'{U}lfar Erlingsson, Vitaly Feldman, Ilya Mironov, Ananth Raghunathan, Kunal
  Talwar, and Abhradeep Thakurta.
\newblock Amplification by shuffling: From local to central differential
  privacy via anonymity.
\newblock In \emph{Proceedings of the Thirtieth Annual ACM-SIAM Symposium on
  Discrete Algorithms}, pp.\  2468–2479, 2019.

\bibitem[Feldman et~al.(2022)Feldman, McMillan, and Talwar]{feldman2022hiding}
Vitaly Feldman, Audra McMillan, and Kunal Talwar.
\newblock Hiding among the clones: A simple and nearly optimal analysis of
  privacy amplification by shuffling.
\newblock In \emph{2021 IEEE 62nd Annual Symposium on Foundations of Computer
  Science (FOCS)}, pp.\  954--964, 2022.
\newblock \doi{10.1109/FOCS52979.2021.00096}.

\bibitem[Ferreira et~al.(2021)Ferreira, Nascimento~da Silva, Gottin, Stelling,
  and Calmon]{ferreira2021bayssgd}
Paulo~Abelha Ferreira, Pablo Nascimento~da Silva, Vinicius Gottin, Roberto
  Stelling, and Tiago Calmon.
\newblock Bayesian sign{SGD} optimizer for federated learning.
\newblock \emph{Advances in Neural Information Processing Systems}, 34, 2021.

\bibitem[Frankle \& Carbin(2018)Frankle and Carbin]{frankle2018lottery}
Jonathan Frankle and Michael Carbin.
\newblock The lottery ticket hypothesis: Finding sparse, trainable neural
  networks.
\newblock In \emph{International Conference on Learning Representations}, 2018.

\bibitem[Girgis et~al.(2021)Girgis, Data, and Diggavi]{girgis2021fl_privacy}
Antonious~M. Girgis, Deepesh Data, and Suhas Diggavi.
\newblock Differentially private federated learning with shuffling and client
  self-sampling.
\newblock In \emph{2021 IEEE International Symposium on Information Theory
  (ISIT)}, pp.\  338--343, 2021.
\newblock \doi{10.1109/ISIT45174.2021.9517906}.

\bibitem[Haddadpour et~al.(2020)Haddadpour, Karimi, Li, and
  Li]{haddadpour2020fedsketch}
Farzin Haddadpour, Belhal Karimi, Ping Li, and Xiaoyun Li.
\newblock Fedsketch: Communication-efficient and private federated learning via
  sketching.
\newblock \emph{arXiv preprint arXiv:2008.04975}, 2020.

\bibitem[Haddadpour et~al.(2021)Haddadpour, Kamani, Mokhtari, and
  Mahdavi]{haddadpour2021federated}
Farzin Haddadpour, Mohammad~Mahdi Kamani, Aryan Mokhtari, and Mehrdad Mahdavi.
\newblock Federated learning with compression: Unified analysis and sharp
  guarantees.
\newblock In \emph{International Conference on Artificial Intelligence and
  Statistics}, pp.\  2350--2358. PMLR, 2021.

\bibitem[Hasircioglu \& Gunduz(2022)Hasircioglu and
  Gunduz]{Hasircioglu2022privacy_participation}
Burak Hasircioglu and Deniz Gunduz.
\newblock Privacy amplification via random participation in federated learning.
\newblock 2022.
\newblock \doi{10.48550/ARXIV.2205.01556}.
\newblock URL \url{https://arxiv.org/abs/2205.01556}.

\bibitem[He et~al.(2015)He, Zhang, Ren, and Sun]{kaiming2015Init}
Kaiming He, Xiangyu Zhang, Shaoqing Ren, and Jian Sun.
\newblock Delving deep into rectifiers: Surpassing human-level performance on
  imagenet classification.
\newblock In \emph{IEEE International Conference on Computer Vision (ICCV)},
  pp.\  1026--1034, 2015.
\newblock \doi{10.1109/ICCV.2015.123}.

\bibitem[He et~al.(2016)He, Zhang, Ren, and Sun]{he2016deep}
Kaiming He, Xiangyu Zhang, Shaoqing Ren, and Jian Sun.
\newblock Deep residual learning for image recognition.
\newblock In \emph{Proceedings of the IEEE conference on computer vision and
  pattern recognition}, pp.\  770--778, 2016.

\bibitem[Hooker et~al.(2020)Hooker, Moorosi, Clark, Bengio, and
  Denton]{hooker2020characterising}
Sara Hooker, Nyalleng Moorosi, Gregory Clark, Samy Bengio, and Emily Denton.
\newblock Characterising bias in compressed models.
\newblock \emph{arXiv preprint arXiv:2010.03058}, 2020.

\bibitem[Imola \& Chaudhuri(2021)Imola and Chaudhuri]{imola2021privacy}
Jacob Imola and Kamalika Chaudhuri.
\newblock Privacy amplification via bernoulli sampling.
\newblock \emph{arXiv preprint arXiv:2105.10594}, 2021.

\bibitem[Isik et~al.(2022)Isik, Weissman, and No]{isik2022information}
Berivan Isik, Tsachy Weissman, and Albert No.
\newblock An information-theoretic justification for model pruning.
\newblock In \emph{International Conference on Artificial Intelligence and
  Statistics}, pp.\  3821--3846. PMLR, 2022.

\bibitem[Ji et~al.(2020)Ji, Jiang, Walid, and Li]{ji2020dynamic}
Shaoxiong Ji, Wenqi Jiang, Anwar Walid, and Xue Li.
\newblock Dynamic sampling and selective masking for communication-efficient
  federated learning.
\newblock \emph{arXiv preprint arXiv:2003.09603}, 2020.

\bibitem[Jiang et~al.(2022)Jiang, Wang, Valls, Ko, Lee, Leung, and
  Tassiulas]{jiang2022model}
Yuang Jiang, Shiqiang Wang, Victor Valls, Bong~Jun Ko, Wei-Han Lee, Kin~K
  Leung, and Leandros Tassiulas.
\newblock Model pruning enables efficient federated learning on edge devices.
\newblock \emph{IEEE Transactions on Neural Networks and Learning Systems},
  April 2022.

\bibitem[Kairouz et~al.(2021)Kairouz, McMahan, Avent, Bellet, Bennis, Bhagoji,
  Bonawitz, Charles, Cormode, Cummings, et~al.]{kairouz2021advances}
Peter Kairouz, H~Brendan McMahan, Brendan Avent, Aur{\'e}lien Bellet, Mehdi
  Bennis, Arjun~Nitin Bhagoji, Kallista Bonawitz, Zachary Charles, Graham
  Cormode, Rachel Cummings, et~al.
\newblock Advances and open problems in federated learning.
\newblock \emph{Foundations and Trends{\textregistered} in Machine Learning},
  14\penalty0 (1--2):\penalty0 1--210, 2021.

\bibitem[Kone{\v{c}}n{\`y} et~al.(2016)Kone{\v{c}}n{\`y}, McMahan, Yu,
  Richt{\'a}rik, Suresh, and Bacon]{konevcny2016federated}
Jakub Kone{\v{c}}n{\`y}, H~Brendan McMahan, Felix~X Yu, Peter Richt{\'a}rik,
  Ananda~Theertha Suresh, and Dave Bacon.
\newblock Federated learning: Strategies for improving communication
  efficiency.
\newblock \emph{arXiv preprint arXiv:1610.05492}, 2016.

\bibitem[Krichevsky \& Trofimov(1981)Krichevsky and
  Trofimov]{krichevsky1981performance}
Raphail Krichevsky and Victor Trofimov.
\newblock The performance of universal encoding.
\newblock \emph{IEEE Transactions on Information Theory}, 27\penalty0
  (2):\penalty0 199--207, 1981.

\bibitem[Krizhevsky et~al.(2009)Krizhevsky, Hinton,
  et~al.]{krizhevsky2009learning}
Alex Krizhevsky, Geoffrey Hinton, et~al.
\newblock Learning multiple layers of features from tiny images.
\newblock 2009.

\bibitem[Li et~al.(2020)Li, Sun, Wang, Duan, Li, Chen, and Li]{li2020lotteryfl}
Ang Li, Jingwei Sun, Binghui Wang, Lin Duan, Sicheng Li, Yiran Chen, and Hai
  Li.
\newblock Lotteryfl: Personalized and communication-efficient federated
  learning with lottery ticket hypothesis on non-iid datasets.
\newblock \emph{arXiv preprint arXiv:2008.03371}, 2020.

\bibitem[Li et~al.(2021)Li, Sun, Zeng, Zhang, Li, and Chen]{li2021fedmask}
Ang Li, Jingwei Sun, Xiao Zeng, Mi~Zhang, Hai Li, and Yiran Chen.
\newblock Fedmask: Joint computation and communication-efficient personalized
  federated learning via heterogeneous masking.
\newblock In \emph{Proceedings of the 19th ACM Conference on Embedded Networked
  Sensor Systems}, pp.\  42--55, 2021.

\bibitem[Lin et~al.(2020)Lin, Wang, Li, Deng, Wang, and Ding]{lin2020esmfl}
Sheng Lin, Chenghong Wang, Hongjia Li, Jieren Deng, Yanzhi Wang, and Caiwen
  Ding.
\newblock Esmfl: Efficient and secure models for federated learning.
\newblock \emph{arXiv preprint arXiv:2009.01867}, 2020.

\bibitem[Lin et~al.(2018)Lin, Han, Mao, Wang, and Dally]{lin2018deep}
Yujun Lin, Song Han, Huizi Mao, Yu~Wang, and Bill Dally.
\newblock Deep gradient compression: Reducing the communication bandwidth for
  distributed training.
\newblock In \emph{International Conference on Learning Representations}, 2018.

\bibitem[Liu et~al.(2021)Liu, Zhao, Zhou, and Xu]{liu2021fedprune}
Yang Liu, Yi~Zhao, Guangmeng Zhou, and Ke~Xu.
\newblock Fedprune: Personalized and communication-efficient federated learning
  on non-iid data.
\newblock In \emph{International Conference on Neural Information Processing},
  pp.\  430--437. Springer, 2021.

\bibitem[McMahan et~al.(2017{\natexlab{a}})McMahan, Moore, Ramage, Hampson, and
  y~Arcas]{mcmahan2017communication}
Brendan McMahan, Eider Moore, Daniel Ramage, Seth Hampson, and Blaise~Aguera
  y~Arcas.
\newblock Communication-efficient learning of deep networks from decentralized
  data.
\newblock In \emph{Artificial intelligence and statistics}, pp.\  1273--1282.
  PMLR, 2017{\natexlab{a}}.

\bibitem[McMahan et~al.(2017{\natexlab{b}})McMahan, Ramage, Talwar, and
  Zhang]{mcmahan2017learning}
H~Brendan McMahan, Daniel Ramage, Kunal Talwar, and Li~Zhang.
\newblock Learning differentially private recurrent language models.
\newblock \emph{arXiv preprint arXiv:1710.06963}, 2017{\natexlab{b}}.

\bibitem[Mironov(2017)]{mironov2017renyi}
Ilya Mironov.
\newblock R{\'e}nyi differential privacy.
\newblock In \emph{IEEE 30th computer security foundations symposium (CSF)},
  pp.\  263--275. IEEE, 2017.

\bibitem[Mitchell et~al.(2022)Mitchell, Ball{\'e}, Charles, and
  Kone{\v{c}}n{\`y}]{mitchell2022optimizing}
Nicole Mitchell, Johannes Ball{\'e}, Zachary Charles, and Jakub
  Kone{\v{c}}n{\`y}.
\newblock Optimizing the communication-accuracy trade-off in federated learning
  with rate-distortion theory.
\newblock \emph{arXiv preprint arXiv:2201.02664}, 2022.

\bibitem[Mohtashami et~al.(2022)Mohtashami, Jaggi, and
  Stich]{mohtashami2022masked}
Amirkeivan Mohtashami, Martin Jaggi, and Sebastian Stich.
\newblock Masked training of neural networks with partial gradients.
\newblock In \emph{International Conference on Artificial Intelligence and
  Statistics}, pp.\  5876--5890. PMLR, 2022.

\bibitem[Mozaffari et~al.(2021)Mozaffari, Shejwalkar, and
  Houmansadr]{mozaffari2021frl}
Hamid Mozaffari, Virat Shejwalkar, and Amir Houmansadr.
\newblock Frl: Federated rank learning.
\newblock \emph{arXiv preprint arXiv:2110.04350}, 2021.

\bibitem[Munir et~al.(2021)Munir, Saeed, Ali, Qazi, and
  Qazi]{munir2021fedprune}
Muhammad~Tahir Munir, Muhammad~Mustansar Saeed, Mahad Ali, Zafar~Ayyub Qazi,
  and Ihsan~Ayyub Qazi.
\newblock Fedprune: Towards inclusive federated learning.
\newblock \emph{arXiv preprint arXiv:2110.14205}, 2021.

\bibitem[Ozfatura et~al.(2021)Ozfatura, Ozfatura, and
  G{\"u}nd{\"u}z]{ozfatura2021time}
Emre Ozfatura, Kerem Ozfatura, and Deniz G{\"u}nd{\"u}z.
\newblock Time-correlated sparsification for communication-efficient federated
  learning.
\newblock In \emph{IEEE International Symposium on Information Theory (ISIT)},
  pp.\  461--466. IEEE, 2021.

\bibitem[Pensia et~al.(2020)Pensia, Rajput, Nagle, Vishwakarma, and
  Papailiopoulos]{pensia2020optimal}
Ankit Pensia, Shashank Rajput, Alliot Nagle, Harit Vishwakarma, and Dimitris
  Papailiopoulos.
\newblock Optimal lottery tickets via subset sum: Logarithmic
  over-parameterization is sufficient.
\newblock \emph{Advances in Neural Information Processing Systems},
  33:\penalty0 2599--2610, 2020.

\bibitem[Ramanujan et~al.(2020)Ramanujan, Wortsman, Kembhavi, Farhadi, and
  Rastegari]{ramanujan2020s}
Vivek Ramanujan, Mitchell Wortsman, Aniruddha Kembhavi, Ali Farhadi, and
  Mohammad Rastegari.
\newblock What's hidden in a randomly weighted neural network?
\newblock In \emph{Proceedings of the IEEE/CVF Conference on Computer Vision
  and Pattern Recognition}, pp.\  11893--11902, 2020.

\bibitem[Reisizadeh et~al.(2020)Reisizadeh, Mokhtari, Hassani, Jadbabaie, and
  Pedarsani]{reisizadeh2020fedpaq}
Amirhossein Reisizadeh, Aryan Mokhtari, Hamed Hassani, Ali Jadbabaie, and
  Ramtin Pedarsani.
\newblock Fedpaq: A communication-efficient federated learning method with
  periodic averaging and quantization.
\newblock In \emph{International Conference on Artificial Intelligence and
  Statistics}, pp.\  2021--2031. PMLR, 2020.

\bibitem[Rissanen \& Langdon(1979)Rissanen and Langdon]{rissanen1979arithmetic}
Jorma Rissanen and Glen~G Langdon.
\newblock Arithmetic coding.
\newblock \emph{IBM Journal of research and development}, 23\penalty0
  (2):\penalty0 149--162, 1979.

\bibitem[Rothchild et~al.(2020)Rothchild, Panda, Ullah, Ivkin, Stoica,
  Braverman, Gonzalez, and Arora]{rothchild2020fetchsgd}
Daniel Rothchild, Ashwinee Panda, Enayat Ullah, Nikita Ivkin, Ion Stoica,
  Vladimir Braverman, Joseph Gonzalez, and Raman Arora.
\newblock Fetchsgd: Communication-efficient federated learning with sketching.
\newblock In \emph{International Conference on Machine Learning}, pp.\
  8253--8265. PMLR, 2020.

\bibitem[Sattler et~al.(2019)Sattler, Wiedemann, M{\"u}ller, and
  Samek]{sattler2019robust}
Felix Sattler, Simon Wiedemann, Klaus-Robert M{\"u}ller, and Wojciech Samek.
\newblock Robust and communication-efficient federated learning from non-iid
  data.
\newblock \emph{IEEE transactions on neural networks and learning systems},
  31\penalty0 (9):\penalty0 3400--3413, 2019.

\bibitem[Seo et~al.(2021)Seo, Ko, Park, Kim, and Bennis]{seo2021communication}
Sejin Seo, Seung-Woo Ko, Jihong Park, Seong-Lyun Kim, and Mehdi Bennis.
\newblock Communication-efficient and personalized federated lottery ticket
  learning.
\newblock In \emph{IEEE 22nd International Workshop on Signal Processing
  Advances in Wireless Communications (SPAWC)}, pp.\  581--585. IEEE, 2021.

\bibitem[Suresh et~al.(2017)Suresh, Felix, Kumar, and
  McMahan]{suresh2017distributed}
Ananda~Theertha Suresh, X~Yu Felix, Sanjiv Kumar, and H~Brendan McMahan.
\newblock Distributed mean estimation with limited communication.
\newblock In \emph{International conference on machine learning}, pp.\
  3329--3337. PMLR, 2017.

\bibitem[Vallapuram et~al.(2022)Vallapuram, Zhou, Kwon, Lee, Xu, and
  Hui]{vallapuram2022hidenseek}
Anish~K Vallapuram, Pengyuan Zhou, Young~D Kwon, Lik~Hang Lee, Hengwei Xu, and
  Pan Hui.
\newblock Hidenseek: Federated lottery ticket via server-side pruning and sign
  supermask.
\newblock \emph{arXiv preprint arXiv:2206.04385}, 2022.

\bibitem[Vargaftik et~al.(2021)Vargaftik, Ben-Basat, Portnoy, Mendelson,
  Ben-Itzhak, and Mitzenmacher]{vargaftik2021drive}
Shay Vargaftik, Ran Ben-Basat, Amit Portnoy, Gal Mendelson, Yaniv Ben-Itzhak,
  and Michael Mitzenmacher.
\newblock Drive: one-bit distributed mean estimation.
\newblock \emph{Advances in Neural Information Processing Systems},
  34:\penalty0 362--377, 2021.

\bibitem[Vargaftik et~al.(2022)Vargaftik, Basat, Portnoy, Mendelson, Itzhak,
  and Mitzenmacher]{vargaftik2022eden}
Shay Vargaftik, Ran~Ben Basat, Amit Portnoy, Gal Mendelson, Yaniv~Ben Itzhak,
  and Michael Mitzenmacher.
\newblock Eden: Communication-efficient and robust distributed mean estimation
  for federated learning.
\newblock In \emph{International Conference on Machine Learning}, pp.\
  21984--22014. PMLR, 2022.

\bibitem[Vogels et~al.(2019)Vogels, Karimireddy, and Jaggi]{vogels2019powersgd}
Thijs Vogels, Sai~Praneeth Karimireddy, and Martin Jaggi.
\newblock Powersgd: Practical low-rank gradient compression for distributed
  optimization.
\newblock \emph{Advances in Neural Information Processing Systems}, 32, 2019.

\bibitem[Wang et~al.(2018)Wang, Sievert, Liu, Charles, Papailiopoulos, and
  Wright]{wang2018atomo}
Hongyi Wang, Scott Sievert, Shengchao Liu, Zachary Charles, Dimitris
  Papailiopoulos, and Stephen Wright.
\newblock Atomo: Communication-efficient learning via atomic sparsification.
\newblock \emph{Advances in Neural Information Processing Systems}, 31, 2018.

\bibitem[Wang et~al.(2019)Wang, Balle, and Kasiviswanathan]{pmlr-v89-wang19b}
Yu-Xiang Wang, Borja Balle, and Shiva~Prasad Kasiviswanathan.
\newblock Subsampled renyi differential privacy and analytical moments
  accountant.
\newblock In Kamalika Chaudhuri and Masashi Sugiyama (eds.), \emph{Proceedings
  of the Twenty-Second International Conference on Artificial Intelligence and
  Statistics}, volume~89 of \emph{Proceedings of Machine Learning Research},
  pp.\  1226--1235. PMLR, 16--18 Apr 2019.

\bibitem[Wen et~al.(2017)Wen, Xu, Yan, Wu, Wang, Chen, and Li]{wen2017terngrad}
Wei Wen, Cong Xu, Feng Yan, Chunpeng Wu, Yandan Wang, Yiran Chen, and Hai Li.
\newblock Terngrad: Ternary gradients to reduce communication in distributed
  deep learning.
\newblock \emph{Advances in neural information processing systems}, 30, 2017.

\bibitem[Yu et~al.(2021)Yu, Nguyen, Anwar, and Jannesari]{yu2021adaptive}
Sixing Yu, Phuong Nguyen, Ali Anwar, and Ali Jannesari.
\newblock Adaptive dynamic pruning for non-iid federated learning.
\newblock \emph{arXiv preprint arXiv:2106.06921}, 2021.

\bibitem[Zhou et~al.(2019)Zhou, Lan, Liu, and Yosinski]{zhou2019deconstructing}
Hattie Zhou, Janice Lan, Rosanne Liu, and Jason Yosinski.
\newblock Deconstructing lottery tickets: Zeros, signs, and the supermask.
\newblock \emph{Advances in neural information processing systems}, 32, 2019.

\end{thebibliography}
\bibliographystyle{iclr2023_conference}

\clearpage
\appendix
\section{\gls{fedpm} Algorithm}
\label{sec:algorithm_app}
We provide the pseudocode for \gls{fedpm} in Algorithms~\ref{algorithm:naive} and~\ref{algorithm:aggregation}. In Algorithm~\ref{algorithm:aggregation}, the prior resetting scheduling policy is controlled by the procedure \emph{ResPrior($t$)}, which may depend on quantities  other than the round number $t$, such as loss.

\begin{algorithm}[!h]
        {\bf Hyperparameters:} learning rate $\eta$, minibatch size $B$, number of local iterations $\tau$.\\
        {\bf Inputs:} local datasets $\mathcal{D}_i$, $i=1, \dots, N$ \\
        {\bf Output:} random seed SEED and binary mask parameters $\bm{m}^{k, T}$\\
            \begin{algorithmic}
            \STATE{At the server, initialize a random network with weight vector $\bm{w}^{\text{init}} \in \mathbb{R}^d$ using a random seed $\mathsf{SEED}$, and broadcast it to the clients. }
            \STATE{At the server, initialize the random score vector $\bm{s}^{g, 0} \in \mathbb{R}^d$, and compute $\bm{\theta}^{g, 0} \gets \text{Sigmoid}(\bm{s}^{g, 0})$.}
            \STATE{At the server, initialize Beta priors $\bm{\alpha}^{g,0} = \bm{\beta}^{g,0} = \bm{\lambda}_0$.}
            \FOR{$t=1, \dots, T$} 
            \STATE{Sample a subset $\mathcal{K}_t \subset \{1, \dots, N\}$ of $|\mathcal{K}_t|=K$ clients without replacement.}
            \STATE{\textbf{On Client Nodes:}}
            \FOR{$k \in \mathcal{K}_t$}
            \STATE{Receive $\bm{\theta}^{g, t-1}$ from the server and set $\bm{s}^{k, t} = \text{Sigmoid}^{-1}(\bm{\theta}^{g, t-1})$.}
            \FOR{$l=1, \dots, \tau$}
            \STATE{$\bm{\theta}^{k, t} \gets \text{Sigmoid}(\bm{s}^{k, t})$ }
            \STATE{Sample binary mask $\bm{m}^{k, t} \sim \text{Bern}(\bm{\theta}^{k, t})$. }
            \STATE{$\dot{\bm{w}}^{k, t} \gets \bm{m}^{k, t} \odot \bm{w}^{\text{init}}$ }
            \STATE{$\text{grad}_{\bm{s}^{k, t}} \gets \frac{1}{B} \sum_{b=1}^{B} \nabla \ell(\dot{\bm{w}}^{k, t}; \mathcal{B}_j^{k})$; $\{ \mathcal{B}_j^{k}\}_{j=1}^{B}$ is uniformly chosen from $\mathcal{D}_k$}
            \STATE{$\bm{s}^{k, t} \gets \bm{s}^{k, t} - \eta \cdot \text{grad}_{\bm{s}^{k, t}}$}
            \ENDFOR \\
            \STATE{$\bm{\theta}^{k, t} \gets \text{Sigmoid}(\bm{s}^{k, t})$}
            \STATE{Sample a binary mask $\bm{m}^{k, t} \sim \text{Bern}(\bm{\theta}^{k, t})$.}
            \STATE{Send the arithmetic coded binary mask $\bm{m}^{k, t}$ to the server.}
            \ENDFOR \\
            \STATE
            \STATE{\textbf{On the Server Node:}}
            \STATE{Receive $\bm{m}^{k, t}$'s from $K$ client nodes.}
            \STATE{$\bm{\theta}^{g, t} = $ BayesAgg( $\{\bm{m}^{k, t}\}_{k \in \mathcal{K}_t}$, $t$) $\quad$ // See Algorithm~\ref{algorithm:aggregation}.}
            \STATE{Broadcast $\bm{\theta}^{g, t}$ to all client nodes.}
            \ENDFOR \\
            \STATE{Sample the final binary mask $\bm{m}^{\text{final}} \sim \text{Bern}(\bm{\theta}^{g, T})$.}
            \STATE{Generate the final model: $\dot{\bm{w}}^{\text{final}} \gets \bm{m}^{\text{final}} \odot \bm{w}^{\text{init}}$.}
            \end{algorithmic}
        \caption{\gls{fedpm}.}
        \label{algorithm:naive}
    \end{algorithm}

\begin{algorithm}[!h]
        {\bf Inputs:} clients' updates $\{\bm{m}^{k, t}\}_{k \in \mathcal{K}_t}$, and round number $t$ \\
        {\bf Output:} global probability mask $\bm{\theta}^{g, t}$\\
        \begin{algorithmic}
        \IF{ResPriors($t$)}
        \STATE{$\bm{\alpha}^{g,t-1} = \bm{\beta}^{g,t-1} = \bm{\lambda}_0 $}
        \ENDIF \\
        \STATE{Compute $\bm{m}^{\text{agg}, t} = \sum_{k \in \mathcal{K}_t} \bm{m}^{k, t}$.}
        \STATE{$\bm{\alpha}^{g, t} = \bm{\alpha}^{g, t-1} + \bm{m}^{\text{agg}, t}$}
        \STATE{$\bm{\beta}^{g, t} = \bm{\beta}^{g, t-1} + K \cdot \bm{1} - \bm{m}^{\text{agg}, t}$}
        \STATE{$\bm{\theta}^{g, t} = \frac{\bm{\alpha}^{g, t} - 1}{\bm{\alpha}^{g, t} + \bm{\beta}^{g, t} - 2}$}
        \STATE{Return $\bm{\theta}^{g, t}$}
        \end{algorithmic}
\caption{BayesAgg.}
        \label{algorithm:aggregation}
\end{algorithm}

\section{Proof of the Upper Bound on the Estimation Error}
\label{sec:error_app}
We now provide proof of the upper bound on the estimation error in Eq.~\ref{eq:error_bound}. Recall that our true mean is $\bm{\bar{\theta}}^{g, t} = \frac{1}{K} \sum_{k \in \mathcal{K}_t} \bm{\theta}^{k, t}$, whereas our estimate is $\bm{\bar{\theta}}^{g, t} = \frac{1}{K} \sum_{k \in \mathcal{K}_t} \bm{m}^{k, t}$, where $\bm{m}^{k, t} \sim \text{Bern}(\bm{\theta}^{k, t})$. Then we can compute the error as

\begin{align}
     \E_{\bm{M}^{k, t} \sim \text{Bern}(\bm{\theta}^{k, t})  \ \forall k \in \mathcal{K}_t} \big  [||\bm{\hat{\bar{\theta}}}^{g, t} - & \bm{\bar{\theta}}^{g, t}||_2^2 \big ] = \sum_{i=1}^d \E_{M_i^{k,t} \sim \text{Bern}(\theta_i^{k, t})  \ \forall k \in \mathcal{K}_t} \left [ \left ( \hat{\bar{\theta}}^{g, t}_i - \bar{\theta}^{g, t}_{i} \right )^2 \right ]\\
     &= \sum_{i=1}^d \E_{M_i^{k,t} \sim \text{Bern}(\theta_i^{k, t})  \ \forall k \in \mathcal{K}_t} \left [ \left ( \frac{1}{K} \sum_{k \in \mathcal{K}_t} (M^{k, t}_i - \theta^{k, t}_{i}) \right )^2 \right ] \\
     &= \frac{1}{K^2} \sum_{i=1}^d \E_{M_i^{k,t} \sim \text{Bern}(\theta_i^{k, t})  \ \forall k \in \mathcal{K}_t} \left [ \left ( \sum_{k \in \mathcal{K}_t} (M^{k, t}_i - \theta^{k, t}_{i}) \right )^2 \right ] \\
    &= \frac{1}{K^2} \sum_{i=1}^d \E_{M_i^{k,t} \sim \text{Bern}(\theta_i^{k, t})  \ \forall k \in \mathcal{K}_t} \left [ \sum_{k \in \mathcal{K}_t} \left ( M^{k, t}_i - \theta^{k, t}_{i}  \right )^2 \right ]\\
    &= \frac{1}{K^2} \sum_{i=1}^d  \sum_{k \in \mathcal{K}_t} \E_{M_i^{k,t} \sim \text{Bern}(\theta_i^{k, t})}  \left [ ( M^{k, t}_i - \theta^{k, t}_{i} )^2 \right ] \\
    &= \frac{1}{K^2} \sum_{i=1}^d  \sum_{k \in \mathcal{K}_t} \left (\E_{M_i^{k,t} \sim \text{Bern}(\theta_i^{k, t})} [  (M^{k, t}_i)^2] - (\theta^{k, t}_{i})^2 \right ) \\
     &= \frac{1}{K^2} \sum_{i=1}^d \sum_{k \in \mathcal{K}_t} \left ( \theta_i^{k, t} - (\theta_i	^{k, t})^2 \right ) \\
     & \leq \frac{d}{4K}.
\end{align}

From (5) to (6), we use the definition of $\hat{\bar{\theta}}_i^{g, t} = \frac{1}{K} \sum_{k=1}^K m^{k, t}_i$ and $\bar{\theta}^{g, t}_{i} = \frac{1}{K} \sum_{k=1}^K \theta^{k, t}_i$. From (7) to (8), we use the fact that $\E_{M_i^{k,t} \sim \text{Bern}(\theta_i^{k,t}) \ \forall k \in \mathcal{K}_t} [M^{k, t}_i - \theta^{k, t}_i] = 0$; and $M^{k, t}_i - \theta^{k, t}_i$ and $M^{l, t}_i - \theta^{l, t}_i$ are independent for $l \neq k \in [K]$. Finally, the inequality in (8) follows from $\theta_i^{k, t} \in [0,1]$ for all $k \in [K]$.

\section{Privacy Amplification and Bias Correction}
\label{sec:privacy_app}

\Gls{dp} guarantees that the probability of an outcome of an algorithm that runs on client data does not change much by a single client's data. This is typically ensured via injecting noise to a function of the client data at a particular step in the algorithm with some utility loss in the application. While there have been many \gls{dp} strategies developed for \gls{fl} and deep learning \citep{abadi2016deep, mcmahan2017learning, agarwal2021skellam, andrew2021differentially}, these strategies typically suffer from severe performance degradation due to noise injection. To make \gls{dp} practical, researchers have explored certain randomization mechanisms that amplify the privacy guarantee. When these mechanisms are parts of the \gls{fl} framework, such as sampling (data \citep{balle2018privacy, pmlr-v89-wang19b} or device \citep{balle2020privacy, girgis2021fl_privacy, Hasircioglu2022privacy_participation}) and shuffling \citep{Erlingsson2019amplification, feldman2022hiding}, the amplification comes for free. This is helpful because the overall process can meet a stronger privacy guarantee without increasing the noise level. 
\gls{fedpm} promises one such amplification due to the stochastic Bernoulli sampling step. We first revisit the definitions of differential privacy \citep{dwork2006calibrating}, R\'enyi divergence, and R\'enyi differential privacy \citep{mironov2017renyi}; and then present the amplification result.

\begin{definition} \textbf{[Adjacent Datasets]}
Two datasets $D, D' \in \mathcal{D}$  are called adjacent if they differ in at most one data sample.
\end{definition}

\begin{definition} \textbf{[$\mathbf{(\epsilon, \delta)}$-DP]}  A randomized mechanism $f: \mathcal{D} \rightarrow \mathcal{R}$ offers $(\epsilon, \delta)$-differential privacy if for any adjacent $D, D' \in \mathcal{D}$ and $\mathcal{S} \subset \mathcal{R}$
\begin{align*}
    \text{Pr}[f(D) \in \mathcal{S}] \leq e^{\epsilon}\text{Pr}[f(D' \in \mathcal{S})] + \delta.
\end{align*}
\end{definition}

\begin{definition} \textbf{[R\'enyi Divergence]} For two probability distributions $P$ and $Q$ defined over $\mathcal{R}$, the R\'enyi divergence of order $\alpha > 1$ is 
\begin{align*}
    D_{\alpha} (P||Q) = \frac{1}{\alpha-1} \log{\E_{x \sim Q} \left ( \frac{P(x)}{Q(x)}\right )^{\alpha} }.
\end{align*}
\end{definition}

\begin{definition} \textbf{[$\mathbf{(\alpha, \epsilon)}$-RDP]} A randomized mechanism $f: \mathcal{D} \rightarrow \mathcal{R}$ offers $\epsilon$-R\'enyi differential privacy of order $\alpha$ (or in short $(\alpha, \epsilon)$-RDP) if for any adjacent $D, D' \in \mathcal{D}$, it holds that
\begin{align*}
    D_{\alpha} (f(D) || f(D')) \leq \epsilon.
\end{align*}
\end{definition}

In particular, \cite{imola2021privacy} have shown that when a sample $\bm{M} \in \{0,1\}^d$ from an already privatized vector $\bm{\theta} \in [c, 1-c]^d$, where $0<c<0.5$, is released to a third party (instead of $\bm{\theta}$ itself), the privacy is amplified under some conditions. More precisely, when there is an $(\alpha, \epsilon)$-R\'enyi Differential Privacy mechanism \citep{mironov2017renyi} that privatizes $\bm{\theta} \in [c, 1-c]^d$, releasing a sample from $\text{Bern}(\bm{\theta})$ yields an improved privacy budget (the smaller $\epsilon$, the better the privacy): $    \epsilon_{amp} \leq \min{\{\epsilon, d \cdot r_{\alpha} (c)\}}$. Here, $r_{\alpha}(p)$ is the binary symmetric R\'enyi divergence function defined as $r_{\alpha}(p) = \frac{1}{\alpha - 1} \log{\left (p^{\alpha} (1-p)^{1-\alpha} + (1-p)^{\alpha} p^{1-\alpha} \right )}$. Notice that \gls{fedpm} already involves this Bernoulli sampling step in the communication protocol and in the forward pass $\bm{m}^{k, t} \sim \text{Bern}(\bm{\theta}^{k, t})$. However, the $d$ term in the upper bound limits the amplification for large model sizes. We believe it is worth exploring a tighter upper bound on $\epsilon_{amp}$ to enjoy privacy amplification in \gls{fedpm} with practical models. Nonetheless, in Appendix~\ref{sec:privacy_app}, we demonstrate the impact of this amplification on a distributed mean estimation problem, described in Figure~\ref{fig:dist_mean_est_diagram}, where the goal is to estimate the true mean of the probability masks $\bar{\bm{\theta}} = \frac{1}{K} \sum_{k=1}^K \bm{\theta}^k$ under communication and privacy constraints. We also provide a bias correction mechanism, specific to our scheme in Figure~\ref{fig:dp_diagram} in Appendix~\ref{sec:privacy_app}, that mitigates the bias due to the DP mechanism and reduces the estimation error. 

Now, suppose that we have an $(\alpha, \epsilon)$-RDP algorithm $f$ that outputs privatized $\bm{\theta}^{k} \in [c, 1-c]^d$ with $0 < c < 0.5$, using local client data $\mathcal{D}_k$. As summarized in Figure~\ref{fig:dp_diagram}, we are interested in what happens when instead of releasing $\bm{\theta}^{k}=f(\mathcal{D}_k)$, the client $k$ releases a Bernoulli sample from it: $\bm{m}^{k} \in \{0,1\}^d \sim \text{Bern}(\bm{\theta}^{k})$. We already explained the advantages in terms of communication bitrate, estimation error, unbiasedness throughout the manuscript; however, this approach also amplifies the privacy guarantees, meaning that it makes the overall privacy budget smaller $\epsilon_{amp} \leq \epsilon$. Quantitatively, \cite{imola2021privacy} showed that after the Bernoulli sampling, the privacy budget of the overall process is

\begin{align*}
    \epsilon_{amp} \leq \min{ \{ \epsilon, d r_{\alpha}(c) \} },
\end{align*}

where $r_{\alpha}(\cdot)$ is the R\'enyi divergence of the binary symmetric function. More precisely, consider $P,Q$ random variables with support on $\{x_1, x_2\} \subset \Theta$ and let $p=\text{Pr}[P=x_1]$, $1-p=\text{Pr}(Q=x_1)$. Then the R\'enyi divergence is defined as

\begin{align*}
    r_{\alpha}(p) = R_{\alpha}(P,Q) = \frac{1}{\alpha-1} \log{(p^{\alpha} (1-p)^{1-\alpha} + (1-p)^{\alpha} p^{1-\alpha})}.
\end{align*}

Notice that \gls{fedpm} already involves this Bernoulli sampling step in the communication protocol and in the forward pass $\bm{m}^{k, t} \sim \text{Bern}(\bm{\theta}^{k, t})$. This implies that \gls{fedpm} improves the privacy guarantee without changing the privacy mechanism -- e.g. without increasing the injected noise level. However, the $d$ term in the upper bound limits the amplification for large model sizes. We believe it is worth exploring a tighter upper bound on $\epsilon_{amp}$ to enjoy privacy amplification in \gls{fedpm} with practical models. Nonetheless, we demonstrate the impact of this amplification on a distributed mean estimation problem, described in Figure~\ref{fig:dp_diagram}, where the probability masks $\bm{\theta}^k \in [c, 1-c]^d$ are a function of client data $\mathcal{D}_k$; and are first corrupted by Gaussian noise, and then clipped to the range $[c, 1-c]^d$. Our goal is, as before, to estimate the true mean $\bm{\bar{\theta}} = \frac{1}{K} \sum_{k \in \mathcal{K}_t}\bm{\theta}^k$ by averaging the sampled binary masks, i.e., $\bm{\hat{\bar{\theta}}} = \frac{1}{K} \sum_{k \in \mathcal{K}_t} \bm{m}^k$. Differently from our previous experiments, we have privacy constraints now, meaning that we want to guarantee $(\epsilon, \delta)$-DP by injecting a Gaussian noise with variance $\sigma^2 = \frac{2 \ln{(1.25 / \delta ) \Delta_2^2}}{\epsilon^2}$ with a small $\epsilon$, where $\delta \approx \frac{1}{N^2}$ and $\Delta_2$ is the $\ell_2$-sensitivity of the probability masks (in our case $\Delta_2 = (1-2c)\sqrt{d}$). We transfer the above amplification results in RDP to DP using the well-known relation:

\begin{remark} \cite{mironov2017renyi} showed that if $f$ is an $(\alpha, \epsilon)$-RDP mechanism, it also satisfies $(\epsilon + \frac{\log{1/\delta}}{\alpha-1}, \delta)$-DP for any $0<\delta<1$.
\end{remark}


 \begin{figure}[h]
 \centering
\includegraphics[width=0.8\textwidth]{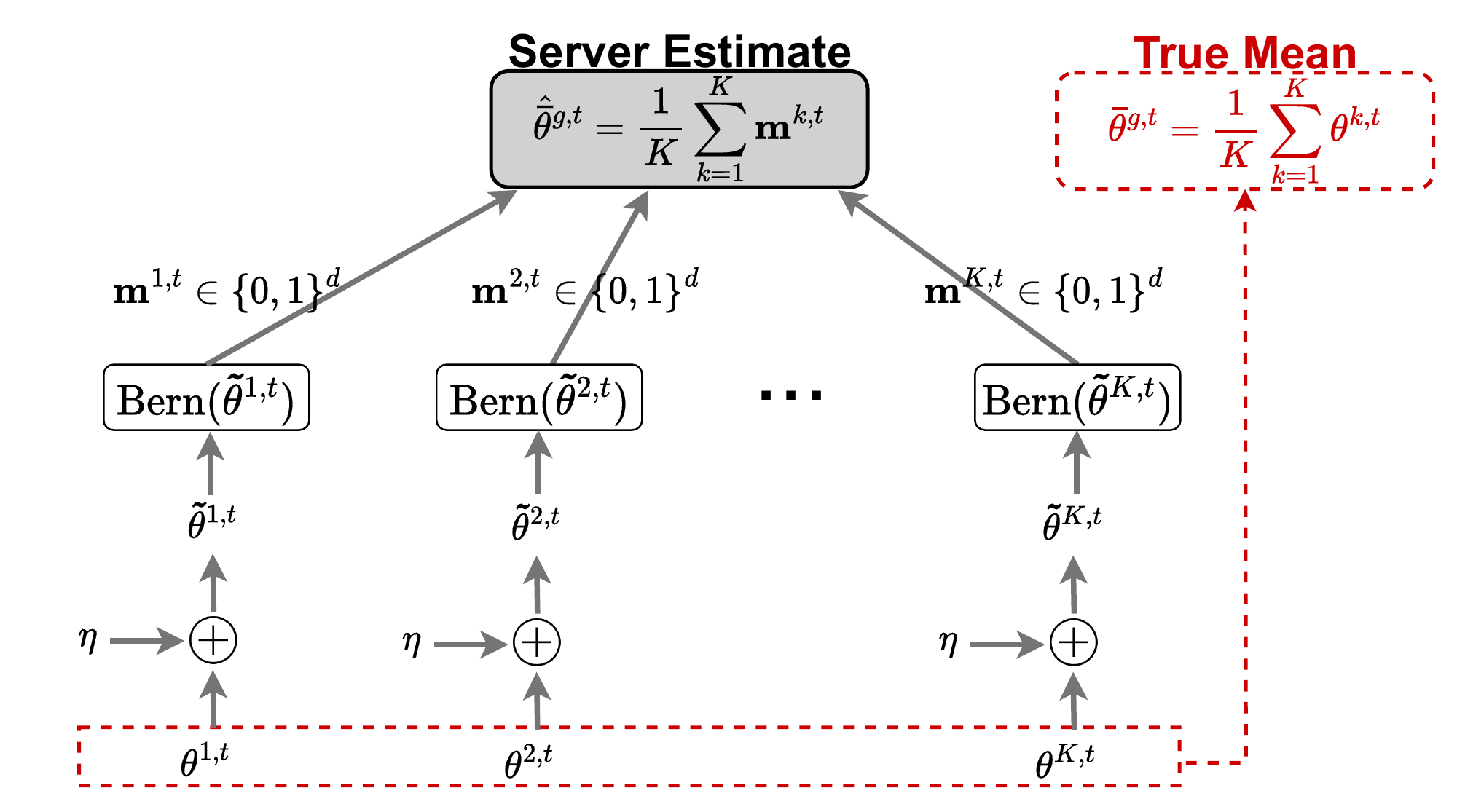}
\caption{Distributed mean estimation scheme in \gls{fedpm}, modified for differential privacy.}\label{fig:dp_diagram}
\end{figure}

Since clipping after the noise addition step would lead to bias in the estimated mean, we work out a bias correction mechanism. We denote with $\theta$ one general parameter at client $k$ for one parameter, with $\tilde{\theta}$ its noisy version, and with $\hat{\theta} = \text{clip}(\tilde{\theta})$ its clipped version. Specifically, if $\tilde{\theta} = \theta + \eta$ is the noisy version of the parameter, where $\eta \sim \mathcal{N}(0, \sigma^2)$, then 
\begin{equation}
\label{eq:clip_eq}
 \text{clip}(\tilde{\theta})=\begin{cases}
    \tilde{\theta}, & \text{if} \quad  c \leq \theta + \eta \leq 1-c\\
    1-c, & \text{if} \quad \theta + \eta  > 1-c \\
    c, & \text{if} \quad \theta + \eta  < c.
  \end{cases}
\end{equation}

We now compute $\mathbb{E}\left[ \hat{M}\right]$, where $\hat{M} \sim \text{Bern}(\hat{\theta})$, to analyze the bias $\mathbb{E}\left[ \hat{M}\right] - \mathbb{E}\left[M\right] = \mathbb{E}\left[ \hat{M}\right] - \theta$, where $M \sim \text{Bern}(\theta)$. First of all, notice that 
\begin{align*}
    \mathbb{E}\left[ \hat{M}\right] &= \int_0^1 \mathbb{E}\left[ \hat{M} | \hat{\theta} = \rho \right] f(\rho) d\rho = \int_0^1 \rho f(\rho) d\rho = \mathbb{E} [ \hat{\theta}].
\end{align*}
And we now compute the mean of the clipped parameter
\begin{align*}
    \label{eq:bias}
    \mathbb{E}\left[ \hat{\theta}\right] &= \int_0^1 \rho f(\rho) d\rho \\
    &= \int_{-\infty}^{+\infty} \text{clip}(\theta + \eta) f(\eta) d\eta  \\
    &= \int_{-\infty}^{c-\theta} c \cdot f(\eta) d\eta + 
    \int_{c-\theta}^{1 -c - \theta} (\theta + \eta) \cdot f(\eta) d\eta + 
     \int_{1 - c -\theta}^{+ \infty} (1-c) \cdot f(\eta) d\eta \\
     &= c \Phi_{\sigma}(c - \theta) + \theta \int_{c-\theta}^{1- c- \theta} f(\eta) d\eta + \int_{c-\theta}^{1-c-\theta} \eta f(\eta) d\eta + (1-c) \left(1 - \Phi_{\sigma}\left(1-c-\theta\right)\right) \\
     &= c \Phi_{\sigma}(c - \theta) +\theta\left[ \Phi_{\sigma}(1-c-\theta) - \Phi_{\sigma}(c-\theta)\right] + \frac{-\sigma}{\sqrt{2\pi}} \left[ e^{\frac{-(1-c-\theta)^2}{2\sigma^2}} - e^{\frac{-(c-\theta)^2}{2\sigma^2}} \right] + \\
     & \quad + (1-c) \left( 1 - \Phi_{\sigma}\left(1-c-\theta \right) \right) \\
     &= 1 -c+ [\theta-1+c] \Phi_{\sigma}(1-c-\theta) + [c - \theta] \Phi_{\sigma}(c-\theta) + \frac{-\sigma e^{\frac{-(c-\theta)^2}{2\sigma^2}}}{\sqrt{2\pi}} \left[ e^{-2(c-\theta) -1} - 1 \right],
\end{align*}

where $\Phi_{\sigma}\left( \cdot \right)$ is the cumulative distribution function of a Gaussian random variable with standard deviation $\sigma$, and zero mean. We use this relation to correct the bias in $\bm{\hat{\bar{\theta}}}$. In practice, to adopt the bias-correction strategy, we sample the function $\mathbb{E}\left[ \hat{\theta}\right]$, which is a function of the true parameter $\theta$, noise standard deviation $\sigma$, and clipping parameter $c$, at $Q$ different points $x_1, \dots, x_Q$, i.e., different values for the uncorrupted $\theta$, and we store the values in a table. Indeed, the values $\sigma$ and $c$ are set at the beginning of the training process, secretly shared among the participants, and never modified. Then, once the server computes an estimate for $\hat{\theta}$, it corrects it by finding the closest outputs of  $\mathbb{E}\left[ \hat{\theta}\right]$ in the stored table, and it inverts the map by choosing the corresponding $x_i$, i.e., the original $\theta$.

\begin{figure}[h]
 \centering
\includegraphics[width=0.7\textwidth]{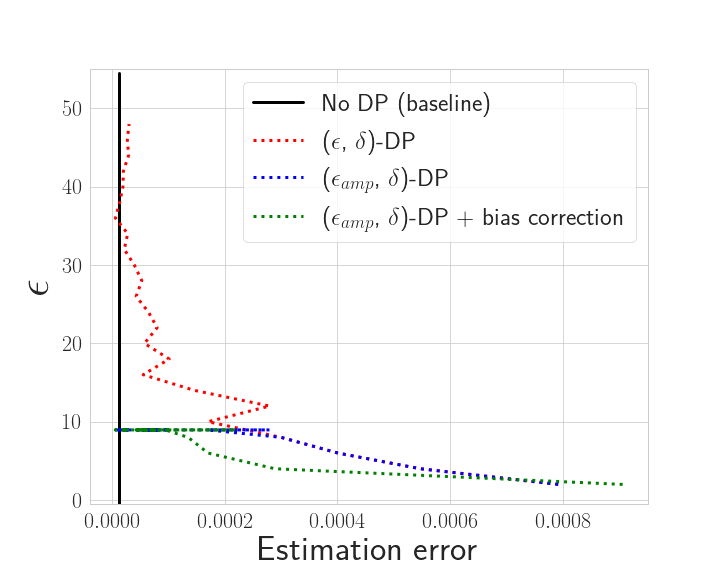}
\caption{The effect of privacy amplification and bias correction in the privacy budget ($\epsilon$) vs. estimation error behavior. Comparing red and blue curves, we see that we can reach small estimation errors without increasing $\epsilon$ thanks to the amplification (see the vertical blue line at low estimation error.). While the red curve and blue curve overlap for $\epsilon < d \cdot r_{\alpha}(c) = 8.96$, in that regime, we benefit from our bias correction strategy to reach a lower error.}\label{fig:DP_exp}
\end{figure}

We conduct our experiments with $N=100$ clients, each having independent probability masks with dimension $d=5$ and range $[0.2, 0.8]$, i.e.,  $\bm{\theta} \in [0.2, 0.8]^5$.  Figure~\ref{fig:DP_exp} shows the estimation error $||\bm{\hat{\bar{\theta}}}^{g, t} - \bm{\bar{\theta}}^{g, t}||_2^2$ under no noise injection case (i.e. no DP) with the black line. Recall that we want to reach a smaller estimation error and smaller $\epsilon$ (i.e., a stronger privacy guarantee). The red curve corresponds to the $\epsilon$ vs. estimation error behavior if Bernoulli sampling did not amplify the privacy. The blue curve shows the amplified $\epsilon$ (i.e. $\epsilon_{amp} \leq \epsilon$) vs. estimation error behavior, and it overlaps with the red curve for  $\epsilon$ values smaller than $d \cdot r_{\alpha}(c) = 8.96$, where there is no privacy amplification, i.e., $\epsilon_{amp} = \epsilon$. However, notice that the blue line never reaches $\epsilon$'s higher than this value due to amplification, while enjoying smaller estimation errors that the red curve can only achieve with very large $\epsilon$. This shows the promise of \gls{fedpm} in having a better privacy-accuracy performance than most baselines that do not have amplification. Finally, the green curve shows that bias correction improves this performance further even with $\epsilon < d \cdot r_{\alpha}(c) = 8.96$ by achieving lower estimation errors with the same $\epsilon$.

\section{Additional Experimental Details}
\label{sec:additional_details_app}
In Table~\ref{tab:architectures}, we provide the architectures for all the models used in our experiments. Clients performed 3 local epochs with a batch size of 128 and a local learning rate of 0.1 in all the experiments. Notice that there is no server learning rate in \gls{fedpm}; instead, we tune the prior resetting schedule in Bayesian aggregation for the experiments in Section~\ref{sec:noniid_exp}. We conducted our experiments on NVIDIA Titan X GPUs on an internal cluster server, using 1 GPU per one run. 

\setlength{\tabcolsep}{3pt}
\begin{table*}[!h]
\centering
\caption{Architectures for $\mathtt{CONV}$-$\mathtt{4}$, $\mathtt{CONV}$-$\mathtt{6}$, and $\mathtt{CONV}$-$\mathtt{10}$ models used in the experiments. 
}
\begin{tabular}{cccc}
\toprule
\textbf{Model}&  $\mathtt{CONV}$-$\mathtt{4}$  & $\mathtt{CONV}$-$\mathtt{6}$   & $\mathtt{CONV}$-$\mathtt{10}$ \\ \midrule
\centered{\\ \\ \\ Convolutional \\ Layers}& \centered{\\ \\ \\ 64, 64, pool \\ 128, 128, pool}
& \centered{\\ \\ 64, 64, pool \\ 128, 128, pool \\ 256, 256, pool } 
& \centered{64, 64, pool \\ 128, 128, pool \\ 256, 256, pool \\ 512, 512, pool \\ 1024, 1024, pool}
\\ \midrule
\centered{Fully-Connected \\ Layers}& \centered{256, 256, 10}
& \centered{256, 256, 10} 
& \centered{256, 256, 100}
\\
\bottomrule
\\
\end{tabular}
\label{tab:architectures}
\end{table*}

In the non-IID and partial participation experiments in Section~\ref{sec:noniid_exp}, to distill the final model, we may apply both stochastic sampling, as during training, or a hard-threshold method, similar to the one adopted in FedMask~\citep{li2021fedmask}. In the latter, a binary mask coefficient $m_i$ is set to $1$ if $\theta_i > \alpha_{\text{ths}}$, and $0$ otherwise. For all experiments but one, when $\alpha_{\text{ths}} \in [0.4, 0.6]$, the thresholding test accuracy is always higher than the sampling method, and so we use the threshold method. However, in the extreme case $c_{\text{max}} = 2$ and $\rho=0.1$, the optimal values for $\alpha_{\text{max}}$ were in $[0.2, 0.4]$ and $[0.6, 0.8]$ in all experiments, probably due to the high randomness given by the highly heterogeneous scenario. Consequently, for the last experiment, we just adopt the stochastic sampling strategy to evaluate the model, as further optimizing the $\alpha_{\text{ths}}$ means adapting to the test dataset, which may corrupt the ability of the model to generalize.

\section{Additional Experimental Results}
\label{sec:additional_results}

\subsection{Additional Experiments on ResNet Architectures}
\label{sec:resnet_app}
In this section, we provide additional experimental results with ResNet-18 \citep{he2016deep} on CIFAR-10 and CIFAR-100 datasets. For these experiments, we focus on IID data distribution and the case when all the clients participate in the training at each round. We use the same hyperparameters from Section~\ref{sec:iid_exp}. We provide the details of the ResNet-18 architecture in Table~\ref{tab:resnet18} below.

\begin{table}[h!]
\caption{ResNet-18 architecture.}
\label{tab:resnet18}
\centering
\begin{tabular}{c|c}
\hline
\textbf{Name}& \textbf{Component}\\
\hline
conv1 & $3\times3$ conv, 64 filters. stride 1, BatchNorm \\
\hline
Residual Block 1 & 
$
\begin{bmatrix}
    3 \times 3 \text{ conv, } 64 \text{ filters} \\
    3 \times 3 \text{ conv, } 64 \text{ filters}
\end{bmatrix}
\times 2$ \\
\hline
Residual Block 2 & 
$
\begin{bmatrix}
    3 \times 3 \text{ conv, } 128 \text{ filters} \\
    3 \times 3 \text{ conv, } 128 \text{ filters}
\end{bmatrix}
\times 2$ \\
\hline
Residual Block 3 & $
\begin{bmatrix}
    3 \times 3 \text{ conv, } 256 \text{ filters} \\
    3 \times 3 \text{ conv, } 256 \text{ filters}
\end{bmatrix}
\times 2$ \\
\hline
Residual Block 4 & $
\begin{bmatrix}
    3 \times 3 \text{ conv, } 512 \text{ filters} \\
    3 \times 3 \text{ conv, } 512 \text{ filters}
\end{bmatrix}
\times 2$ \\
\hline
Output Layer & $4 \times 4$ average pool stride 1, fully-connected, softmax \\
\hline
\end{tabular}
\end{table}

 \begin{figure}[!h]
 \centering
\includegraphics[width=.40\textwidth]{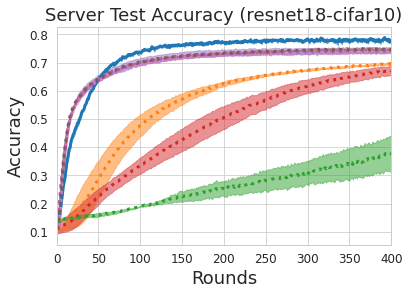}
\includegraphics[width=.58\textwidth]{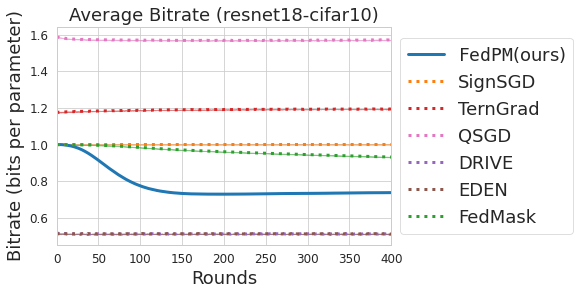}

\caption{Accuracy and bitrate comparison of \gls{fedpm} with baselines SignSGD~\citep{bernstein2018signsgd}, TernGrad~\citep{wen2017terngrad}, QSGD~\citep{alistarh2017qsgd}, DRIVE~\citep{vargaftik2021drive}, EDEN~\citep{vargaftik2022eden}, and FedMask~\citep{li2021fedmask}, with ResNet-18 on CIFAR-10.}\label{fig:resnet_cifar10}
\end{figure}

 \begin{figure}[!h]
 \centering
\includegraphics[width=.40\textwidth]{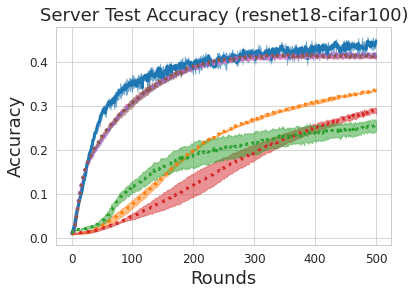}
\includegraphics[width=.58\textwidth]{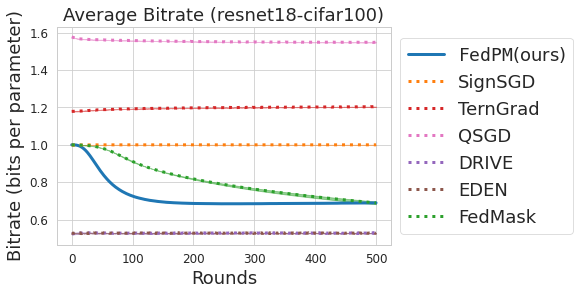}

\caption{Accuracy and bitrate comparison of \gls{fedpm} with baselines SignSGD~\citep{bernstein2018signsgd}, TernGrad~\citep{wen2017terngrad}, QSGD~\citep{alistarh2017qsgd}, DRIVE~\citep{vargaftik2021drive}, EDEN~\citep{vargaftik2022eden}, and FedMask~\citep{li2021fedmask}, with ResNet-18 on CIFAR-100.}\label{fig:resnet_cifar100}
\end{figure}

Figures~\ref{fig:resnet_cifar10} and~\ref{fig:resnet_cifar100} show the results on CIFAR-10 and CIFAR-100 datasets, respectively. It is seen that \gls{fedpm} outperforms all the baselines in terms of accuracy. Although DRIVE and EDEN require approximately 0.1 smaller bitrates than \gls{fedpm}, they also reach lower accuracy. In summary, the advantages of \gls{fedpm} discussed in the main manuscript carry over to ResNet-18 model as well.

\subsection{Bitrate Considerations on non-iid Data}
\label{sec:bitrate_results}

We now report the communication bitrate considerations on the non-IID data split experiments described in Section~\ref{sec:noniid_exp}. Table~\ref{tab:non_iid_4_class_rate} reports the average bitrate needed by different algorithms over the whole training process when $c_{\text{max}} = 4$ and $c_{\text{max}} = 2$. By simply multiplying the obtained average bitrate by the total number of rounds $t_{\text{max}} = 200$, we obtain the total number of bits one element in the global probability mask needs to converge to its final value, indicating the total amount of information communicated during the training process.

We first observe that both DRIVE and EDEN consume almost the same amount of bits no matter the system configuration and round number (very small variance), and it is instead model dependent (see Figure~\ref{fig:comparison_iid}). On the contrary, \gls{fedpm} and QSGD report higher bitrate variability, as it depends on both the training phase and system setting. As already observed in Section~\ref{sec:iid_exp}, FedMask balances almost uniformly the binary updates, leading to a bitrate that is basically fixed to $1$. For both $c_{\text{max}} = 4$ and $c_{\text{max}} = 2$, \gls{fedpm} yields the smallest bitrate when $\rho = 1$, whereas for the other scenarios, EDEN and DRIVE are slightly more efficient. We argue that this is motivated by the fact that, as the learning task becomes harder due to the high system heterogeneity, all the models struggle to converge to good and stable solutions, which means that \gls{fedpm} is still uncertain about the \textit{weights' importance probabilities} $\bm{\theta}$, setting many of them close to $0.5$. However, we think that this may be a useful feature of \gls{fedpm} to quantify its internal uncertainty, which we will further analyze.

\begin{table}[h!]
\centering
\resizebox{\columnwidth}{!}{
\begin{tabular}{ cccccc }
 \toprule
& \textbf{Algorithm} & $\bm{\rho = 1}$& $\bm{\rho = 0.5}$ & $\bm{\rho = 0.2}$ & $\bm{\rho = 0.1}$ \\
 \midrule
& DRIVE~\citep{vargaftik2021drive} & $0.885 \pm 9\cdot 10^{-5}$  & $\bm{0.885 \pm 1\cdot 10^{-4}}$ & $ \bm{0.885 \pm 6 \cdot 10^{-5}}$ & $\bm{0.885 \pm 1\cdot 10^{-4}}$ \\
& EDEN~\citep{vargaftik2022eden} & $0.885 \pm 1\cdot 10^{-4}$  & $\bm{0.885 \pm 1\cdot 10^{-4}}$ & $ \bm{0.885 \pm 8\cdot 10^{-5}}$ & $\bm{0.885 \pm 1\cdot 10^{-4}}$ \\
$c_{\text{max}} = 4$ & QSGD~\citep{alistarh2017qsgd} & $0.982 \pm 0.027$  & $0.923 \pm 0.029$ & $ 1.188 \pm 0.034 $ & $0.910 \pm 0.05$ \\
& FedMask~\citep{li2021fedmask}& $1.000 \pm 3 \cdot 10^{-6}$ & $1.000 \pm 8 \cdot 10^{-8}$ & $ 1.000 \pm 2\cdot 10^{-6}$ & $1.000 \pm 6 \cdot 10^{-7}$ \\
& \gls{fedpm} (Ours) & $\bm{0.863 \pm 0.077}$ & $0.912 \pm 0.056$ & $ 0.965 \pm 1\cdot 0.01812$ & $0.996 \pm 0.003$ \\
 \midrule
&  DRIVE~\citep{vargaftik2021drive} & $0.885 \pm 7\cdot 10^{-5}$ & $\bm{0.885 \pm 2\cdot 10^{-4}}$ & $ \bm{0.885 \pm 7\cdot 10^{-5}}$ & $\bm{0.885 \pm 2\cdot 10^{-4}}$ \\
& EDEN~\citep{vargaftik2022eden} & $0.885 \pm 1\cdot 10^{-4}$ & $\bm{0.885 \pm 7\cdot 10^{-5}}$ & $ \bm{0.885 \pm 6\cdot 10^{-5}}$ & $\bm{0.885 \pm 7\cdot 10^{-5}}$ \\
$c_{\text{max}} = 2$ & QSGD~\citep{alistarh2017qsgd} & $1.230 \pm 0.043$ & $1.234 \pm 0.038$ & $ 1.100 \pm 0.01$ & $1.082 \pm 0.01 $ \\
& FedMask~\citep{li2021fedmask}& $1.000 \pm 2 \cdot 10^{-6}$ & $1.000 \pm 2 \cdot 10^{-6}$ & $ 1.000 \pm 1\cdot 10^{-5}$ & $1.000 \pm 2 \cdot 10^{-7}$ \\
& \gls{fedpm} (Ours) & $\bm{0.868 \pm 0.076}$ & $0.904 \pm 0.063$ & $ 0.980 \pm 0.014$ & $0.997 \pm 0.01$ \\
\bottomrule
\end{tabular}}
\caption{Average bitrate $\pm \sigma$ over the whole training process in non-IID data split with $c_{\text{max}} = 4$ and $c_{\text{max}} = 2$, and partial participation with ratios $\rho = \{0.1, 0.5, 1\}$, for \gls{fedpm}, FedMask, and the strongest baselines in the IID experiments: EDEN, DRIVE, and QSGD. The training duration was set to $t_{\text{max}} = 200$ rounds.}
\label{tab:non_iid_4_class_rate}
\end{table}


To conclude the analysis, we also report the \gls{fedpm} bpp for the final model, which is an indication of the average number of bits needed per one parameter of the model. In the case of $c_{\text{max}} = 4$, the final model sizes are $0.79$ bpp, $0.834$ bpp, and $0.99$ bpp, when $\rho = \{0.1, 0.5, 1\}$, respectively. When $c_{\text{max}} = 2$, the final model sizes are $0.8$ bpp, $0.817$ bpp, and $0.992$ bpp. Consequently, at the end of the training process, \gls{fedpm} remains the most efficient option, as already observed in Section~\ref{sec:iid_exp}.

\end{document}